\pdfoutput=1

\documentclass[11pt]{article}

\usepackage[preprint]{acl}
\usepackage{times}
\usepackage{latexsym}
\usepackage[T1]{fontenc}
\usepackage[utf8]{inputenc}

\usepackage{microtype}

\usepackage{inconsolata}

\usepackage{amsmath}

\definecolor{mydarkblue}{RGB}{4, 91, 191}
\definecolor{darkred}{RGB}{181, 53, 16}
\newcommand{\gr}[1]{{\color{gray}#1}}

\usepackage{booktabs}
\usepackage{caption}
\usepackage{amssymb}
\usepackage{bm}

\usepackage{tikz,pgfplots,pgfplotstable}
\usetikzlibrary{tikzmark}
\usepgfplotslibrary{colorbrewer}
\pgfplotsset{compat=1.11}
\pgfplotscreateplotcyclelist{customcols}{darkred, darkred!40, gray!40, gray!20, mydarkblue!40, mydarkblue}
\pgfplotscreateplotcyclelist{blueonly}{mydarkblue}
\newlength{\plotwidth}
\newlength{\plotheight}
\newcommand{\labelsize}[1]{\fontsize{8}{6}\selectfont#1}
\newcommand{\means}[2]{\textit{#1}=#2}

\usepackage{adjustbox}
\usepackage[inline]{enumitem}

\newcommand{{\perc}}{\kern1pt\%}
\newcommand{\fslash}{/\allowbreak{}}

\usepackage[german, english]{babel}

\newcommand{\german}[1]{\foreignlanguage{german}{#1}}
\newcommand{\translation}[1]{\textcolor{gray}{\textit{#1}}}
\newcommand{\surveynote}[1]{\textit{#1}}
\newcounter{questionnr}
\newcommand{\question}[2]{
\refstepcounter{questionnr}\label{#2}
\vspace{\baselineskip}
\noindent\textbf{\thequestionnr.~\german{#1}}
}
\newcommand{\translq}[1]{\translation{#1}}
\newcommand{\sectionboundary}{\begin{center}
\textcolor{mydarkblue!50}{\rule{0.1\columnwidth}{1pt}}
\end{center}}

\usepackage{tcolorbox}
\newcommand{\definition}[2]{%
\begin{tcolorbox}[colback=white,colframe=mydarkblue!50]%
\noindent\german{#1}
\translation{#2}
\end{tcolorbox}}

\title{What Do Dialect Speakers Want?\\ A Survey of Attitudes Towards Language Technology for German Dialects}

\usepackage{fontawesome5}
\newcommand{\lmu}{\faMountain}
\newcommand{\mcml}{\faRobot}
\newcommand{\itu}{\faCompass}
\newcommand{\unilux}{\faComment} %

\usepackage{stackengine}
\newcommand{\lmuMcml}{\kern1pt\stackon[1pt]{\scriptsize\mcml}{\scriptsize\lmu}}
\newcommand{\lmuMcmlItu}{\kern1pt\stackon[1pt]{\scriptsize\mcml\kern1pt\itu}{\scriptsize\lmu}}
\newcommand{\lux}{\kern1pt\textsuperscript{\scriptsize\unilux}}

\author{Verena Blaschke{\lmuMcml} \enspace{} Christoph Purschke{\lux} \enspace{} Hinrich Schütze{\lmuMcml} \enspace{} Barbara Plank{\lmuMcmlItu}\\
\textsuperscript{\lmu} Center for Information and Language Processing (CIS), LMU Munich, Germany \\
\textsuperscript{\mcml} Munich Center for Machine Learning (MCML), Munich, Germany \\
\textsuperscript{\unilux} Department of Humanities, University of Luxembourg, Luxembourg \\
\textsuperscript{\itu} Department of Computer Science, IT University of Copenhagen, Denmark  \\
{\tt \{verena.blaschke, b.plank\}@lmu.de}}

\newcommand{\myrho}{$\rho$}
\newcommand{\mysigma}{$\sigma$}
\newcommand{\mymu}{$\mu$}
\newcommand{\mygeq}{$\geq$}

\begin{document}
\maketitle
\begin{abstract}
Natural language processing (NLP) has largely focused on modelling standardized languages. 
More recently, attention has increasingly shifted to local, non-standardized languages and dialects. However, the relevant speaker populations' needs and wishes with respect to NLP tools are largely unknown.  
In this paper, we focus on dialects and regional languages related to German -- a group of varieties that is heterogeneous in terms of prestige and standardization.
We survey speakers of these varieties (\textit{N}=327) and present their opinions on hypothetical language technologies for their dialects. 
Although attitudes vary among subgroups of our respondents, we find that respondents are especially in favour of potential NLP tools that work with dialectal input (especially audio input) such as virtual assistants, 
and less so for applications that produce dialectal output such as machine translation or spellcheckers. 
\end{abstract}

\section{Introduction}
\label{sec:intro}
Most natural language processing (NLP) research focuses on languages with many speakers, high degrees of standardization and large amounts of available data \citep{joshi-etal-2020-state}.
Only recently, more NLP projects have started to include local, non-standardized languages and dialects. 
However, different speakers and cultures have different needs. 
As recently echoed by multiple researchers, 
the creation of language technologies (LTs) should take into account what the relevant speaker community finds useful
\citep{bird-2020-decolonising, bird-2022-local, liu-etal-2022-always, mukhija2021social-good},
and communities can differ from one another in that regard \citep{lent-etal-2022-creole}.

In this work, we focus on dialects and regional languages\footnote{%
Our survey also includes responses by speakers of Low German, which is officially recognized as a regional language.
}
closely related to German (for the sake of simplicity, we use `dialects' to refer to these varieties in this paper).
With dialect competence generally being in decline in the German-speaking area, today, dialect speakers usually also speak %
Standard German, 
while dialects often are replaced by regiolects -- intermediate varieties between standard and dialect 
\citep{kehrein2019vertical}. 
Speaker attitudes towards dialects vary greatly \citep[pp.~155--167]{gaertig2010spracheinstellungen}.

\begin{figure}
    \centering
    \includegraphics[width=\columnwidth, trim={0pt 16mm 0pt 4mm}, clip]{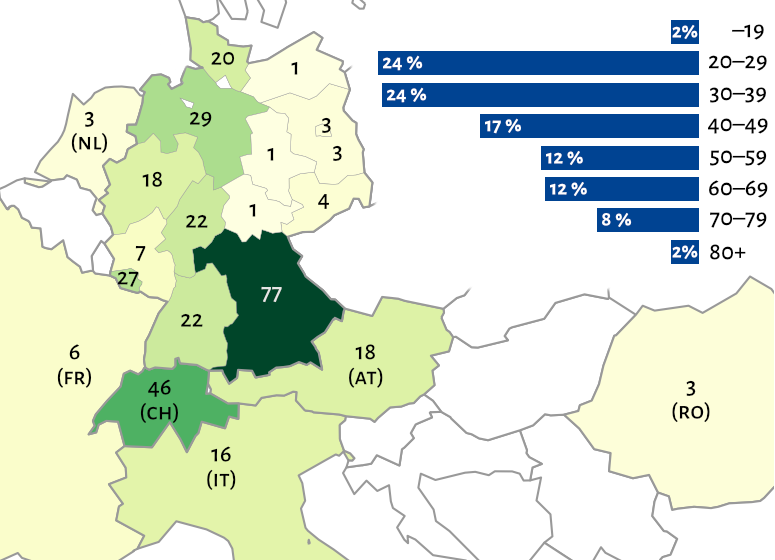}
    \caption{\textbf{Countries and German states in which the respondents' dialects are spoken}, with the number of respective respondents, and the overall age distribution.} %
    \label{fig:map}
\end{figure}

Although these dialects are predominantly spoken and only few of them have traditional orthographies, many of them are also used in written, digital contexts 
\citep{androutsopoulos2003onlinegemeinschaften}.
Accordingly, some NLP datasets based (primarily) on such digital data exist, and a small number is also annotated for NLP tasks \citep{blaschke-etal-2023-survey}.
Several recent publications feature LTs for German dialects, such as 
machine translation \citep{haddow-etal-2013-corpus, honnet-etal-2018-machine, lambrecht-etal-2022-machine, aepli-etal-2023-benchmark, her2024investigating}, 
speech-to-text
\citep{herms-etal-2016-corpus, nigmatulina-etal-2020-asr, gerlach-etal-2022-producing}
and text-to-speech systems \citep{gutscher2023austrian-tts}, 
and slot and intent detection for conversational assistants 
\citep{van-der-goot-etal-2021-masked, aepli-etal-2023-findings, winkler-etal-2024-slot-intent, abboud-oz-2024-towards-equitable}.

To investigate what speaker communities are interested in, we survey dialect speakers from different German-speaking areas
(Figure~\ref{fig:map}).
We gather a snapshot of their current attitudes towards LTs to answer the following questions:
\begin{enumerate*} 
[label=\itshape{}Q\arabic*)]
    \item Which dialect technologies do respondents find especially useful (\S\ref{sec:results-lts})?
    \item Does this depend on whether the in- or output is dialectal, and on whether the LT works with speech or text data 
    (\S\ref{sec:results-inputoutput})?
    \item How
    does this 
    reflect
    relevant     sociolinguistic factors (\S\ref{sec:results-groups})? %
\end{enumerate*}

\section{Related Work}
\label{sec:related}

\label{sec:surveys-digital}

The closest survey to ours on investigating attitudes of speakers of non-standard language varieties towards LTs is by \citet{lent-etal-2022-creole}. 
They conducted a survey on the actual and desired LT use by speakers of different creoles (\textit{N}=37). 
They find that the needs vary from speaker community to speaker community, and that speakers who are also highly proficient in the local high-prestige language are less interested in creole LTs.
Of the technologies included in the survey, speech-related technologies (%
transcription and synthesis) are the most popular; machine translation (MT) and question answering software are also desired by multiple communities, 
while spellcheckers are controversial.

\citet{soria-etal-2018-dldp} surveyed speakers of four regional European languages%
\footnote{Karelian (\textit{N}=156, \citealp{dldp2017karelian}), 
Breton (\textit{N}=202, \citealp{dldp2017breton}), 
Basque (\textit{N}=427, \citealp{dldp2017basque}), 
Sardinian (\textit{N}=516,  \citealp{dldp2017sardinian}).}
about whether and why they use (or do not use) 
their languages in digital contexts.
When asked about the desirability of currently unavailable spellcheckers and MT systems, more respondents judged both %
as desirable than %
not,
although the exact proportions vary 
across communities.
\citeauthor{millour2019getting} (\citeyear{millour2019getting};  \citeyear{millour2020myriadisation}, pp.~230, 239) found similar results in surveys of 
Mauritian Creole (\textit{N}=144) and Alsatian speakers (\textit{N}=1,224).

Conversely, \citet{way2022report} investigate actual LT use by speakers of different European national languages (91--922 respondents per country).
The most commonly used LTs are MT, search engines and spell- or grammar checkers.
When respondents do not use specific LTs, %
this can simply be due to 
the absence of such technologies for certain languages, but also 
due to
a lack of interest in specific language--LT combinations.

Recently, several surveys have also investigated speaker community perspectives regarding LTs for many different indigenous language communities \citep{mager-etal-2023-ethical, cooper-etal-2024-things, dolinska-etal-2024-akha, tonja2024nlp}.
However, these surveys focus on languages with very different socio\-linguistic contexts than the ones in our survey and that are unrelated to their respective local high-resource languages. %

\section{Questionnaire}

Our questionnaire is aimed at speakers of German dialects and related regional languages and consists of two main parts:
We ask our participants about their dialect, and we ask about their opinions on hypothetical LTs for their dialect.
Several of the questions regarding dialect use %
are inspired by %
\citet{soria-etal-2018-dldp} and \citet{millour2020myriadisation}, and
we choose a similar selection of LTs as \citet{way2022report} (%
\S\ref{sec:results-lts}).
For each technology, 
we provide a brief definition to make the survey accessible to a broad audience (e.g., `Speech-to-text systems transcribe spoken language. They are for instance used for automatically generating subtitles or in the context of dictation software.'). 
We then ask participants to rate on a 5-point Likert scale how useful they would find such a tool for their dialect.
We allow respondents to elaborate on their answers in comment fields.
The full questionnaire is in Appendix~\S\ref{sec:full-questionnaire}. %

The questionnaire was written in German, and was estimated to take between 10--15~minutes for completion.\footnote{80\perc{} of our respondents took <15~min to fill out the entire questionnaire, and 60\perc{} even less than 10~min.} 
It was online for three weeks in September and October 2023 and got disseminated via word of mouth, social media, mailing lists and by contacting dialect and heritage societies.
Our results are hence based on a convenience sample.

\section{Results}

We reached 441 people, 327 of whom are self-reported dialect speakers and finished the entire questionnaire -- their responses are presented in the following.
Detailed answer distributions are in Appendix~\S\ref{sec:full-questionnaire}; correlations are in \S\ref{sec:correlations}.

\subsection{Dialect Background and Attitudes}
\label{sec:results-dialect}

Most of our respondents answer that they have a very good command of their dialect (68{\perc}) and acquired it as a mother tongue (71{\perc}).
Figure~\ref{fig:map} %
shows where the respondents' dialects are spoken (and their age distribution): mostly in Germany (72{\perc}), %
followed by Switzerland (14{\perc}) and Austria (6{\perc}).%
\footnote{The other varieties are spoken in areas with minority speaker communities: Italy (Bavarian), France (Alemannic), 
the Netherlands (Low German), and
Romania.
The geographic distribution of our respondents is not representative of the overall dialect speaker population.
}
Nearly a quarter (24\perc{}) each are in their twenties and thirties, almost all others are older. %
When rating how traditional their dialect is on a scale from 1 (traditional dialect that people from other regions have trouble understanding) to 5 (regionally marked German easily understood by outsiders), the largest group of respondents (35\perc) indicated a~2 (\mymu=2.6, \mysigma=1.1).

Just over half of our respondents (52{\perc}) speak their dialect on a daily basis, and 43\perc{} indicate that they would like to use their dialect in all areas of life.
Most respondents (70\perc) value the diversity of their dialect.
Nearly two thirds (65\perc) are opposed to having a standardized orthography for their dialect.
Just over half of the respondents (53\perc) say that their dialect is only spoken and not well-suited for written communication -- nevertheless, two thirds (66{\perc})
also write their dialect, even if rarely.
Many (63\perc) find it easy to read dialectal texts written by others. 
Written dialect is commonly used for communicating with others -- the most common writing scenarios are text messages (57{\perc}, multiple responses possible), followed by letters\fslash{}emails (26{\perc}), social media posts and comments (19{\perc} each) -- but also for notes to oneself including diary entries (19{\perc}).

About a third (35\perc) indicate that they are actively engaged in dialect preservation pursuits (multiple responses possible): 13{\perc} as members of dialect preservation societies, 4{\perc} as teachers, and 22{\perc} %
in other ways.
Write-in comments by the last group point out other language-related professions, but also include speaking the dialect in public or with children as a means of active dialect preservation.\footnote{%
The write-in answers contain 13~mentions of speaking the dialect with family members (especially children and grandchildren), 18~mentions of simply speaking the dialect (in public), 14~mentions of carrying out dialect-related research (as a job or a hobby), and 10~mentions of using the dialect in the context of literature or music, with slight overlap between these groups. 
None of these subgroups are concentrated in any specific area, but instead include respondents from areas where dialects and regional languages have very different statuses (cf.\ \S\ref{sec:results-groups}): Low German speakers as well as other German respondents, Swiss respondents, respondents from countries where German is a minority language, and so on.} 
We compare the opinions of respondents with and without such dialect engagement in \S\ref{sec:results-groups}.

14\perc{} 
of our respondents are familiar with at least one LT that already caters to their dialect.
Just over half of the respondents (54\perc) indicate that their dialect being represented by more LTs would make it more attractive to younger generations,
and a smaller group (31\perc) says they would use their dialect more often given suitable LTs.

\subsection{Which dialect LTs are deemed useful?}
\label{sec:results-lts}

Figure~\ref{fig:opinions-lt} shows our respondents' opinions on LTs (\textit{Q1}), and Appendix~\S\ref{sec:ranking} presents the average scores per LT when responses are mapped to a numerical scale.
While there are diverging opinions on every LT -- there is no single technology that (nearly) all respondents consider useful or useless for their dialect -- some trends emerge, %
as we discuss next.\footnote{Of the technologies we included, MT, search engines and spell checkers are the most used LTs in the EU \citep[p.~26]{way2022report}. We assume that the tools that people use a lot are also tools they generally deem useful, yet those tools are all ranked relatively low in our results – suggesting that our results reveal attitudes on dialect LTs rather than LTs in general.}
Overall, the responses are generally correlated with each other: respondents who think positively\fslash{}negatively of one technology tend to think similarly about others.
Nevertheless, some LTs are overall more popular, and some less so:

\pgfplotstableread[col sep=comma]{fig/opinions_lt_all.csv}\opiniondata

\setlength{\plotheight}{9cm}
\setlength{\plotwidth}{\columnwidth}

\begin{figure}[t]
\resizebox{\columnwidth}{!}{
\setlength{\lineskiplimit}{-\maxdimen} %
\begin{tikzpicture}[font=\sffamily]
\begin{axis}[
    width=\plotwidth, height=\plotheight,
    xbar stacked,
    xmin=0, xmax=100,
    axis line style={draw=none},
    ytick=data,
    yticklabels from table={\opiniondata}{LT},
    yticklabel style={align=right},
    enlarge y limits={abs=4mm},
    ytick style={draw=none},
    tick label style={font=\labelsize},
    xtick={0, 20, 40, 60, 80, 100},
    xticklabels={0, 20, 40, \gr{60}, \gr{80}, \gr{100\,\%}},
    grid=major, tick align=inside,
    major grid style={draw=gray!30},
    x tick style={color=gray!30},
    legend style={
        legend columns=2,
        transpose legend,
        at={(xticklabel cs:0.43)},
        anchor=north,
        draw=none,
        /tikz/every even column/.append style={column sep=0.25cm},
        font=\labelsize,
        },
    legend cell align={left},
    cycle list name=customcols,
    every axis plot/.style={fill},
    nodes near coords style={font=\labelsize, /pgf/number format/assume math mode, /pgf/number format/.cd, fixed, precision=0},
    nodes near coords, 
    ]
\addplot+[text=black!15] table[x=No, meta=LT,y expr=\thisrow{YPos}] {\opiniondata};
\addplot+[text=black!5] table[x=RatherNo, meta=LT,y expr=\thisrow{YPos}] {\opiniondata};
\addplot+[text=black!70] table[x=CannotJudge, meta=LT,y expr=\thisrow{YPos}] {\opiniondata};
\addplot+[text=black!50] table[x=NeitherNor, meta=LT,y expr=\thisrow{YPos}] {\opiniondata};
\addplot+[text=black!5] table[x=RatherYes, meta=LT,y expr=\thisrow{YPos}] {\opiniondata};
\addplot+[text=black!15] table[x=Yes, meta=LT,y expr=\thisrow{YPos}] {\opiniondata};
\legend{Useless, Rather useless, Cannot judge, Neither/nor, Rather useful, Useful}
\end{axis}
\begin{axis}[
    width=\plotwidth, height=\plotheight,
    axis line style={draw=none},
    xmin=0, xmax=100,
    xtick={0, 20, 40, 60, 80, 100},
    xticklabels={, \gr{100}, \gr{80}, \gr{60}, 40, 20, 0\,\%},
    axis x line*=top,
    ytick=none,
    yticklabels={},
    axis y line*=right,
    y tick style={opacity=0},
    x tick style={color=gray!30},
    tick label style={font=\labelsize},
    ]
\end{axis}
\end{tikzpicture}
}
\caption{%
\textbf{Opinions on potential language technologies for dialects.}
\means{STT}{speech-to-text}, \means{TTS}{text-to-speech}, 
\means{dial}{dialect}, \means{deu}{German}, \means{oth}{other languages}, 
\means{MT}{machine translation}, %
\means{cannot judge}{skip question}.
}
\label{fig:opinions-lt}
\end{figure}

\paragraph{Virtual assistants and chatbots}
The most clearly favoured LT by dialect speakers in our survey are virtual assistants (such as Siri or Alexa) that can respond to dialectal input (71\perc{} in favour, 20\perc{} against).
Chatbots that can handle dialectal input are less popular, but still deemed useful by a slight majority (52\perc{}).
Systems that could output dialectal responses are less popular: 48\perc{} would find virtual assistants that answer in dialect useful, and 34\perc{} think so about chatbots. %

\paragraph{Speech-to-text and text-to-speech}
When asked about %
speech-to-text (STT) software, a majority (61\perc) is in favour of systems that transcribe spoken dialect into written Standard German, and a slightly smaller majority (58\perc) is in favour of written dialectal output.
When it comes to text-to-speech (TTS) systems that synthesize dialect text into a spoken form, the respondents are even more split, with 47\perc{} in favour and 35\perc{} against.

\paragraph{Machine translation}
We ask for opinions on four different configurations regarding automatic translation of written texts: each possible combination for translation into vs.\ out of the dialect and from\fslash{}into Standard German vs.\ a foreign language.
All options are to some degree controversial among the respondents, with translation from the dialect into Standard German being the most popular (52\perc{} in favour) and from the dialect into a foreign language the least popular (25\perc{} in favour).

\paragraph{Search engines}
Search engines that could deal with dialectal input are controversial, with 43\perc{} each in favour of and against this LT, although the negative group holds stronger opinions.
Some write-in comments question whether (monolingual) information retrieval would produce useful results or mention finding it easier to write in Standard German rather than in a dialect, but others voice a desire to be able to find results for queries including dialectal terms with no direct German equivalent.

\paragraph{Spellcheckers}
Most respondents (59\perc) are opposed to %
spell- or grammar checkers
for their dialect, although a quarter (25\perc) is in favour.
Several respondents mention opposition to spellcheckers since they want their dialectal writing to exactly reflect the pronunciation and word choices of their local dialect and would be bothered if a spellchecker changed them to an arbitrary standardized version of the dialect.

\subsection{Are there differences for dialect input vs.\ output and text vs.\ speech?}
\label{sec:results-inputoutput}

As seen in the previous section, there is a general tendency to prefer versions of LTs that process dialectal input rather than produce dialectal output (\textit{Q2}).
Several write-in comments voice worries about dialectal output not modelling their dialect accurately enough.
Additionally, technologies dealing with spoken language tend to be rated more positively than those focusing on text only.

\paragraph{Correlation with attitudes towards orthography}
Being in favour of a standardized dialect orthography is positively, albeit not very strongly, correlated with being in favour of any technology involving a written version of the dialect and\fslash{}or (written or spoken) dialectal output (Spearman's \myrho{} values between 0.14 and 0.47 per LT with \textit{p}-values <0.001).%

\subsection{Do attitudes reflect sociolinguistic factors?}
\label{sec:results-groups}

To address \textit{Q3} and the heterogeneity of our respondent group, we compare answers between larger subgroups.
We summarize the results of \textit{t-}tests with \textit{p}-values <0.05. %
Appendix~\S\ref{sec:subgroup-comparisons} provides more details, together with two additional comparisons that only have small effect sizes (speaker age and dialect traditionality).

\paragraph{Language activists}
Since language activists might have overly enthusiastic attitudes compared to the speaker population at large \citep{soria-etal-2018-dldp},
we compare those who report involvement in dialect preservation (`activists', \textit{N=}115) to those who do not (\textit{N=}212).
Activists are generally more in favour of LTs for dialects, with statistically significant differences for (any kind of) machine translation, TTS software, spellcheckers, and search engines, as well as for written dialect output options for STT, chatbots and virtual assistants.
Removing the activists' responses from our analysis only barely changes the order of preferred LTs~(\S\ref{sec:ranking}).

\paragraph{Region}

Additionally, we compare
three large regional subgroups with different sociolinguistic realities.
In Germany and Austria, traditional dialects have been partially replaced by more standard-like regiolects, while dialects have high prestige in Switzerland 
where Standard German is often reserved for writing
\citep{kehrein2019vertical, ender2009stellenwert}.
Low German, traditionally spoken in parts of Northern Germany and the Eastern Netherlands, is officially recognized as a language and is more distantly related to Standard German than the other varieties our participants speak.
Its speaker numbers are in decline, but many Northern Germans think Low German should receive more support in, e.g., public schools \citep{adler2016niederdeutsch}.

We compare the opinions of Swiss (\textit{N}=46) and Low German (\textit{N}=58)  respondents to German and Austrian non-Low-German speakers (\textit{N}=191).\footnote{We identify the Low German respondents based on the dialect they indicated speaking~(\S{}A\ref{q:which-dialect}), combined with region information for respondents who supplied ambiguous dialect names. For the (other) German, Austrian, and Swiss respondents, we used region information.}
Our Low German respondents are more in favour of a standardized orthography and of spellcheckers than our other German and Austrian respondents, the Swiss respondents less so.
This is unsurprising in that several orthographies have been proposed for %
Low German, whereas (typically spoken) dialects and (typically written) Standard German exist in a diglossic state in Switzerland.
Nevertheless, \textit{both} groups are more in favour of STT software with dialectal output. %
The Low German respondents are more in favour of chatbots with dialectal answers, TTS, (any kind of) MT and search engines.
Swiss Germans rate virtual assistants that can handle dialectal input as more desirable (%
87\perc{} in favour), and are more in favour of STT software with Standard German output.

\section{Discussion and Conclusion}

We surveyed speakers of dialect varieties on their attitudes towards LTs. Generally, 
the survey participants prefer LTs working with dialect input rather than output.
They also tend to prefer tools that process speech over those for text~(\textit{Q2}).
This is consistent with \citeauthor{chrupala-2023-putting}'s (\citeyear{chrupala-2023-putting}) argument that NLP should focus more on spoken language to better represent actual language use. 
It also reflects the complex, often conflicting attitudes speakers of multiple varieties have towards competing linguistic and social norms. %
Consequently, the most popular potential dialect LTs~(\textit{Q1}) in our survey process spoken dialectal input: virtual assistants %
with dialect input and speech-to-text systems.

However, like \citet{lent-etal-2022-creole}, we find that different speaker communities vary in their attitudes towards LTs~(\textit{Q3}). For instance, opinions on the standardization of a dialect are a relevant factor regarding the desirability of written LTs. 
Nevertheless, the acceptance and rejection of LTs is related to individual factors beyond just attitudes, e.g., experience with and trust in digital technology.

We hope that our study inspires other NLP researchers to actively consider the wants and needs of the relevant speaker communities.
Based on the results of this study, we also encourage the dialect NLP community to pursue more work on spoken language processing.

\section*{Ethical Considerations}

We only collected responses from participants who consented to having their data saved and analyzed for research purposes.
We did not ask for personally identifying information.
We store responses on a local server and only share results based on aggregate analyses.
Appendix~\S\ref{sec:full-questionnaire} contains the full questionnaire including the introduction where we describe the purpose of the study and explain what data we collect and how we use the data.

Participation was completely anonymous, voluntary and with no external reward.
We do not see any particular risks associated with this work.

\section*{Limitations}

Our results are based on a convenience sample; neither the geographic or age distribution are representative of the population at large (dialect-speaking or not).
Language activists are over-represented (hence our additional analysis in \S\ref{sec:results-groups} and Appendices \S\ref{sec:ranking} and~\S\ref{sec:subgroup-comparisons}), and participating in the survey may have been especially of interest to people who feel (in one way or another) strongly about the topic of dialects and technology.
Even so, our respondents are not a monolith in their opinions and we can see meaningful differences between the relative popularity of different technologies.

We aimed to keep participation effort low and therefore limited the number of questions we included.
We considered asking ``Would you use \textit{X} if it existed?''\ in addition to ``Would you find \textit{X} useful?''\ to explicitly disentangle the participants' own needs from what are possibly the perceived needs of the community. 
We decided against this in order to keep the questionnaire as short as possible and because we were unsure how accurate such assessment would be.

The scale of our answer possibilities uses ``not useful'' as the opposite of ``useful.'' 
However, it would be interesting to instead use a scale from ``harmful'' to ``useful'' in future surveys, in order to get a better impression of whether respondents who deem an LT useless in our version of the survey find it actively harmful or merely uninteresting.

To minimize the total time needed to fill out the questionnaire and to guarantee the privacy of the respondents after asking respondents about what specific dialect they speak (used later to identify the Low German speakers), we intentionally kept additional demographic questions at a minimum and did not ask about education, income, gender, or similar variables.

As this survey is based on self-reporting, we expect discrepancies between reported and actual opinions and behaviour. %
Since participation was anonymous and entirely voluntary with no external reward, we think it unlikely for participants to lie about their opinions. It is likely, though, that (especially younger) participants overstate their dialect competence or the traditionality of their dialect, in line with overall dialect dynamics in German \citep{purschke2011regionalsprache}.

\section*{Acknowledgements}

We thank everybody who participated in or shared the survey.
We also thank Yves Scherrer and Frauke Kreuter for their advice, and the MaiNLP and CIS members as well as the anonymous reviewers for their feedback.

This research is supported by the ERC Consolidator Grant DIALECT 101043235. %
We also gratefully acknowledge partial funding by the European Research Council (ERC \#740516). %

\bibliography{bibliography}

\appendix

\section{Questionnaire}
\label{sec:full-questionnaire}

\raggedbottom
\newlength\baritem
\setlength{\baritem}{0.4pt}

\newcommand{\countbar}[1]{%
\textcolor{mydarkblue}{\rule[-1mm]{#1\baritem}{3.6mm}}
\ifnum0#1>29
    \ifnum0#1<100
        \hspace{-5mm}{\scriptsize \textcolor{white}{#1}}
    \else
        \hspace{-7mm}{\scriptsize \textcolor{white}{#1}}
    \fi
\else
    {\scriptsize \textcolor{mydarkblue}{#1}}
\fi
}

\newenvironment{answerbars}
{
\vspace{0.5\baselineskip}
\noindent
\begin{tabular}{@{}>{\raggedleft\arraybackslash}p{103pt}@{\hspace{4pt}}p{112pt}@{}}}
{\end{tabular}}

\newcommand{\descr}[2]{\labelsize \begin{tabular}[c]{@{}r@{}}#1\\ #2\end{tabular}}

\newcommand{\entry}[3]{\descr{#1}{#2} & \countbar{#3} \\[5pt]}
\newcommand{\onelineentry}[2]{{\labelsize #1} & \countbar{#2} \\[5pt]}

\newcommand{\asterisknote}[1]{\fontsize{8}{10}\selectfont #1}

In this section, we reproduce the questions and answers from our survey, in the original wording as well as in translation.
Translations are in \translation{grey italics,} remarks about the questionnaire are in \textit{black italics.}
Answer options that end with a colon~(:) came with an optional text input field in the questionnaire.
All questions except for the first two could be skipped without answering.

\sectionboundary

\noindent
\german{Herzlich willkommen, servus, grüezi \& moin!}\par
\german{\textbf{Sprachtechnologie ist momentan allgegenwärtig,} ob bei
Übersetzungsprogrammen, Chatbots oder anderen Anwendungen. Hauptsächlich
unterstützen diese Anwendungen lediglich Standardsprachen -- was nicht unbedingt
dem entspricht, wie wir im Alltag Sprache verwenden.}\par
\german{Daher möchten wir herausfinden, wie Sie als Sprecher*innen von Dialekten und
Regionalsprachen möglicher Sprachtechnologie für Ihre Sprachform
gegenüberstehen: \textbf{welche Anwendungen halten Sie für wünschenswert bzw.
unnötig?}}\\
\translation{Welcome and hello [in different dialects]!}\par 
\translation{\textbf{Language technology is currently omnipresent} -- be it in the context of translation software, chatbots or other applications. Such applications primarily support standard languages -- which is not necessarily how we use language in our everyday lives.}\par
\translation{Because of this we would like to find out what you as speakers of dialects and regional languages think of potential technologies for your language variety: \textbf{which applications do you find desirable or useless?}}

\sectionboundary

\noindent
\german{Das Ausfüllen des Fragebogens dauert etwa 10--15 Minuten.}\par
\german{Wir behandeln Ihre Antworten vertraulich und veröffentlichen diese nur in anonymisierter Form und ohne dass Rückschlüsse auf Ihre Person gezogen werden können.}\par
\german{Genauere Details:}\par
\german{Ziel der Befragung ist es zum einen, herauszufinden, ob es Unterschiede zwischen den Arten von Sprachtechnologien gibt, die
Dialektsprecher*innen tendenziell als nützlich bzw. nutzlos bewerten.
Zum anderen möchten wir herausfinden, ob ein statistischer Zusammenhang zwischen diesen Antworten und dem Dialekthintergrund und -gebrauch der Befragten besteht.}\par
\german{Die Antworten werden auf einem Server der LMU in München
gespeichert. Wir speichern dabei nur Ihre Antworten und den
Antwortzeitpunkt (um die typische Ausfülldauer besser einzuschätzen),
nicht aber Ihre IP-Adresse.}
\german{Wir geben die Daten \textit{nicht} an Dritte weiter, sondern veröffentlichen
lediglich Ergebnisse, die auf Aggregatdaten und statistischen Analysen beruhen. Zudem zitieren wir gegebenenfalls aus (optional gegebenen)
Kommentarfeld-Antworten.}\par
\german{Kontaktmöglichkeit bei Fragen oder Kommentaren zu dieser Umfrage:}
\surveynote{[Contact data of first author].}\par
\german{Vielen Dank für Ihre Teilnahme!}\\
\translation{This questionnaire takes about 10--15 minutes to fill out.}\par
\translation{We treat your answers as confidential and only share them as anonymized data that do not allow drawing any inferences about your identity.}\par
\translation{More detailed information:}\par
\translation{The goal of this survey is firstly to determine whether there are differences between the types of language technologies that dialect speakers tend to find useful or useless.
Additionally, we would like to find out whether there is a statistical correlation between the answers and the dialect background of the participants.}\par
\translation{We store the answers on an LMU server in Munich. This only includes storing your answers and the time of the questions are answered (to better estimate the typical response duration), but not your IP address. We do {\normalfont not} share your data with third parties, but only share results based on aggregated data and statistical analyses. Additionally, we might cite (optional) write-in answers from comment fields.}\par
\translation{Contact person in case of questions about or comments on this study: [Contact data of first author].}\par
\translation{Thank you very much for participating!}

\begin{itemize}
    \item[$\square$] \german{Ich stimme zu, dass meine Antworten wie oben beschrieben zu Forschungszwecken gespeichert und ausgewertet werden.} \translation{I consent to my answers being stored and analyzed for research purposes as outlined above.}
\end{itemize}
\surveynote{The survey only progresses if this box is checked.}

\sectionboundary

\noindent
In dieser Umfrage untersuchen wir Sprachen und Sprachformen, die sich deutlich vom Hochdeutschen unterscheiden. Damit meinen wir mit dem Deutschen verwandte \textbf{Regionalsprachen} sowie \textbf{Dialekte, Mundarten} und \textbf{Platt}-Varianten, die meist für eine kleine Region typisch sind und von Außenstehenden nicht ohne Weiteres verstanden werden können. Ein paar Beispiele dafür sind das Eifeler Platt, Allgäuerisch, Bairisch oder Nordfriesisch. \textbf{Der Einfachheit halber verwenden wir im Folgenden „Dialekt“ als Sammelbegriff für all diese Sprachformen.}
\translation{In this survey, we focus on languages and language varieties that are clearly distinct from Standard German. To be precise, we are interested in \textbf{regional languages} related to German as well as \textbf{dialects}\footnote{\textit{In the German version, we include different terms that all translate to ``dialect'' but are used in different regions.}} that usually are typical for a small region and cannot easily be understood by outsiders. Some examples are Eifelplatt, Allgäu dialects, Bavarian and North Frisian. \textbf{For the sake of simplicity, we will use ``dialect'' as umbrella term for all of these language varieties in the following.}}\par

\surveynote{This introduction is partially based on the one from the REDE project surveys \citep{regionalsprachede}.}

\question{Können Sie einen deutschen Dialekt sprechen?}{q:skills}
\translq{Can you speak a German dialect?}

\begin{answerbars}
\entry{Ja, sehr sicher}{\translation{Yes, very well}}{223}
\entry{Ja, gut}{\translation{Yes, well}}{71}
\entry{Ein wenig}{\translation{A bit}}{33}
\entry{Nein}{\translation{No}}{16}
\end{answerbars}

\noindent
\surveynote{The 16 respondents who answered `no' are excluded from the analysis. The survey automatically ended for them, showing the message:}
``Alle weiteren Fragen richten sich nur an SprecherInnen eines deutschen Dialekts bzw. einer mit dem Deutschen nahe verwandten Regionalsprache. Vielen Dank für Ihre Teilnahme!''
\translation{``All further questions are only for speakers of a German dialect or a regional language closely related to German. Thank you for participating!''}

\question{Um welchen Dialekt handelt es sich?}{q:which-dialect}
\translq{Which dialect specifically?}

\surveynote{327 write-in answers.}

\question{Wann haben Sie diesen Dialekt gelernt?}{q:aoa}
\translq{When did you learn this dialect?}

\begin{answerbars}
\entry{Als Muttersprache}{\translation{As mother tongue}}{231}
\entry{Kindheit/Jugend}{\translation{Childhood/youth}}{84}
\entry{Später}{\translation{Later}}{12}
\end{answerbars}

\question{In welchem Land befindet sich der Ort, an dem Ihr Dialekt gesprochen wird (z.B.
Ihr Heimatort)?}{q:where}
\translq{In which country is the location where your dialect is spoken (e.g., your hometown)?}

\surveynote{See Figure~\ref{fig:map}.}

\filbreak
\question{In welchem Bundesland befindet sich dieser Ort?}{q:where-deu}
\translq{In which German state is this location?}

\surveynote{Only asked if the previous answer is `Germany'. See Figure~\ref{fig:map}.}

\question{Wie sehr entspricht Ihr Dialekt dem traditionellen Dialekt des Ortes?}{q:traditionality}
\translq{How much does your dialect resemble the traditional dialect of this location?}

\begin{answerbars}
\entry{1 -- Mein Dialekt ist}{sehr traditionell...*}{57}
\onelineentry{2}{114}
\onelineentry{3}{93}
\onelineentry{4}{38}
\entry{5 -- Mein Dialekt ist eher}{regional gefärbtes...**}{25}
\end{answerbars}

{\asterisknote
*\german{1 -- Mein Dialekt ist sehr traditionell und für Außenstehende aus anderen Regionen sehr schwer zu verstehen.}
\translation{1~--~My dialect is very traditional and very hard to understand for outsiders from other regions.}\par
**\german{5 -- Mein Dialekt ist eher regional gefärbtes Deutsch, das auch von Außenstehenden recht einfach verstanden wird.}
\translation{5~--~My dialect is more like regionally marked German that is relatively easily understood by outsiders.}\par
}

\question{Wie häufig sprechen Sie Ihren Dialekt?}{q:freq}
\translq{How often do you speak your dialect?}

\surveynote{The answer options are based on those in the surveys summarized by \citet{soria-etal-2018-dldp}.}

\begin{answerbars}
\entry{Täglich}{\translation{Daily}}{170}
\entry{Mehrmals pro Woche}{\translation{Multiple times per week}}{48}
\entry{Min. einmal pro Woche}{\translation{At least once a week}}{32}
\entry{Min. einmal pro Monat}{\translation{At least once a month}}{29}
\entry{Seltener}{\translation{More rarely}}{46}
\entry{Nie}{\translation{Never}}{2}
\end{answerbars}

\question{Schreiben Sie manchmal Ihren Dialekt?}{q:writing}
\translq{Do you ever write your dialect?}

\surveynote{This question and the next one are modelled after questions by \citet[pp.~228, 237--238]{millour2020myriadisation}.}

\begin{answerbars}
\entry{Ja (egal ob häufig o. selten)}{\translation{Yes (whether often or rarely)}}{217}
\entry{Nein, ich weiß nicht wie}{\translation{No, I don't know how}}{14}
\entry{Nein, ich habe dazu}{keine Gelegenheit*}{26}
\entry{Nein, mein Dialekt ist eine}{gesprochene Sprachform...**}{62}
\entry{Nein, aus anderen Gründen:}{\translation{No, for other reasons:}}{8}
\end{answerbars}

{\asterisknote
*\translation{No, I don't have any opportunity for this}\par

**\german{Nein, mein Dialekt ist eine gesprochene Sprachform und ich möchte ihn nicht schreiben}
\translation{No, my dialect is a spoken form of language and I don't want to write it}\par
}

\question{Was schreiben Sie in Ihrem Dialekt? {\normalfont (Mehrfachantworten möglich)}}{q:write-what}
\translq{What do you write in your dialect? (Multiple answers possible)}

\surveynote{Only asked if previous is `yes'. 217 participants responded:}

\begin{answerbars}
\entry{Nachrichten in}{Chatprogrammen...*}{187}
\entry{Briefe, Emails}{\translation{Letters, emails}}{86}
\entry{Einträge auf sozialen Medien}{\translation{Social media posts}}{63}
\entry{Kommentare auf soz. Med.}{\translation{Social media comments}}{61}
\entry{Sachtexte, z.B. als}{Blogposts...**}{19}
\entry{Prosa, Poesie}{\translation{Prose, poetry}}{43}
\entry{Witze}{\translation{Jokes}}{24}
\entry{Rezepte}{\translation{Recipes}}{6}
\entry{Notizen an mich selbst,}{Tagebucheinträge***}{63}
\entry{Andere/weitere Sachen:}{\translation{Other/additional things:}}{22}
\end{answerbars}

{\asterisknote
*\german{Nachrichten in Chatprogrammen, Messengern (wie WhatsApp), SMS} 
\translation{Texts in messaging apps (like WhatsApp), text messages}\par

**\german{Sachtexte, z.B. als Blogposts oder auf Wikipedia}
\translation{Non-fiction texts, e.g., blog posts or Wikipedia articles}\par

***\translation{Notes to myself, diary entries}\par
}

\question{Setzen Sie sich aktiv für den Erhalt Ihres Dialekts ein? {\normalfont (Mehrfachantworten möglich)}}{q:activism}
\translq{Are you actively involved in preserving your dialect? (Multiple answers possible)}

\surveynote{This question is based on questions in the surveys by \citet{soria-etal-2018-dldp} and \citet[pp.~227, 235]{millour2020myriadisation}. 323 respondents answered:}

\begin{answerbars}
\entry{Ja, in einem Verein}{zur Mundartpflege*}{43}
\entry{Ja, als Lehrer*in}{\translation{Yes, as a teacher}}{14}
\entry{Ja, anderweitig:}{\translation{Yes, in another way:}}{73}
\entry{Nein}{\translation{No:}}{212}
\end{answerbars}

{\asterisknote
*\translation{Yes, in a dialect preservation society}\par
}

\question{Wie alt sind Sie?}{q:age}
\translq{How old are you?}

\begin{answerbars}
\entry{19 Jahre oder jünger}{\translation{19 years or younger}}{7}
\entry{20--29 Jahre}{\translation{20--29 years}}{79}
\entry{30--39 Jahre}{\translation{30--39 years}}{78}
\entry{40--49 Jahre}{\translation{40--49 years}}{54}
\entry{50--59 Jahre}{\translation{50--59 years}}{39}
\entry{60--69 Jahre}{\translation{60--69 years}}{38}
\entry{70--79 Jahre}{\translation{70--79 years}}{25}
\entry{80 Jahre oder älter}{\translation{80 years or older}}{7}
\end{answerbars}

\question{Weitere Kommentare zu Ihrem Dialekt oder zu den vorherigen Fragen: {\normalfont(Optional)}}{q:comments-dial}
\translq{Additional comments on your dialect or the preceding questions: (Optional)}

\surveynote{61 write-in answers.}

\sectionboundary

\noindent
\german{In diesem Abschnitt fragen wir Sie zu Ihrer Meinung zu verschiedenen dialektbezogenen
Themen. \textbf{Dabei gibt es keine richtigen\fslash{}falschen oder erwünschten\fslash{}unerwünschten
Antworten,} sondern wir sind an Ihrer persönlichen Meinung interessiert.}
\translation{In this section we ask you about your opinion on different dialect-related topics. \textbf{There are no right/wrong or desired/undesired answers,} we are simply interested in your personal opinion.}

\question{Stimmen Sie den folgenden Aussagen zu?}{q:opinions-dialect}
\translq{Do you agree with the following statements?}

\surveynote{Statements presented in a randomized order:}

\begin{itemize}
    \item 
\german{Die Vielfalt der unterschiedlichen Ausprägungen meines Dialekts ist eine Stärke.}
\translation{The diversity of the different varieties of my dialect is a strength.}

    \item 
\german{Mein Dialekt ist in erster Linie eine gesprochene Sprachform und nicht für die schriftliche Kommunikation geeignet.}
\translation{My dialect is primarily a spoken form of language and not suited for written communication.}
\surveynote{This question is based on an answer option in the survey by \citet[pp.~228, 237]{millour2020myriadisation} (see also question~\ref{q:writing} in this appendix).}

    \item 
\german{Ich möchte meinen Dialekt in allen Lebensbereichen verwenden.}
\translation{I'd like to be able to use my dialect in any aspect of life.}
\surveynote{This question is based on a question by \citet{soria-etal-2018-dldp} and \citet[pp.~229, 239]{millour2020myriadisation}.} %

    \item 
\german{Wenn ich einen Text lese, den jemand anderes in meinem Dialekt verfasst hat, fällt es mir schwer, ihn zu verstehen.}
\translation{When I read text that someone else wrote in my dialect, I have trouble understanding it.}

    \item 
\german{Es sollte eine standardisierte Rechtschreibung für meinen Dialekt geben.}
\translation{There should be a standardized orthography for my dialect.}
\end{itemize}

\filbreak
\noindent\surveynote{Answer options:}
\begin{itemize}
    \item Ja, auf jeden Fall \translation{Yes, absolutely}
    \item Eher ja \translation{Rather yes}
    \item Weder noch \translation{Neither/nor}
    \item Eher nein \translation{Rather no}
    \item Nein, gar nicht \translation{Absolutely not}
    \item Keine Angabe \translation{Prefer not to say}
\end{itemize}

\noindent\surveynote{The answer distributions (in \%) are as follows:}
\pgfplotscreateplotcyclelist{customcols}{darkred, darkred!40, gray!40, gray!20, mydarkblue!40, mydarkblue}

\pgfplotstableread[col sep=comma]{fig/opinions_dialect.csv}\opiniondata

\setlength{\plotheight}{5cm}
\setlength{\plotwidth}{0.95\columnwidth}

\begin{tikzpicture}[font=\sffamily]
\begin{axis}[
    width=\plotwidth, height=\plotheight,
    xbar stacked,
    xmin=0, xmax=100,
    axis line style={draw=none},
    ytick=data,
    yticklabels from table={\opiniondata}{LT},
    yticklabel style={align=right},
    enlarge y limits={abs=4mm},
    ytick style={draw=none},
    tick label style={font=\labelsize},
    xtick={0, 20, 40, 60, 80, 100},
    xticklabels={0, 20, 40, \gr{60}, \gr{80}, \gr{100\,\%}},
    grid=major, tick align=inside,
    major grid style={draw=gray!30},
    x tick style={color=gray!30},
    legend style={
        legend columns=2,
        transpose legend,
        at={(xticklabel cs:0.4)},
        anchor=north,
        draw=none,
        /tikz/every even column/.append style={column sep=0.25cm},
        font=\labelsize,
        },
    legend cell align={left},
    cycle list name=customcols,
    every axis plot/.style={fill},
    nodes near coords style={font=\labelsize, /pgf/number format/assume math mode, /pgf/number format/.cd, fixed, precision=0},
    nodes near coords, 
    ]
\addplot+[text=black!15] table[x=No, meta=LT,y expr=\thisrow{YPos}] {\opiniondata};
\addplot+[text=black!5] table[x=RatherNo, meta=LT,y expr=\thisrow{YPos}] {\opiniondata};
\addplot+[text=black!70] table[x=CannotJudge, meta=LT,y expr=\thisrow{YPos}] {\opiniondata};
\addplot+[text=black!50] table[x=NeitherNor, meta=LT,y expr=\thisrow{YPos}] {\opiniondata};
\addplot+[text=black!5] table[x=RatherYes, meta=LT,y expr=\thisrow{YPos}] {\opiniondata};
\addplot+[text=black!15] table[x=Yes, meta=LT,y expr=\thisrow{YPos}] {\opiniondata};
\legend{Disagree, Dis.\ somewhat, N/A, Neither/nor, A.\ somewhat, Agree}
\end{axis}
\begin{axis}[
    width=\plotwidth, height=\plotheight,
    axis line style={draw=none},
    xmin=0, xmax=100,
    xtick={0, 20, 40, 60, 80, 100},
    xticklabels={, \gr{100}, \gr{80}, \gr{60}, 40, 20, 0\,\%},
    axis x line*=top,
    ytick=none,
    yticklabels={},
    y tick style={opacity=0},
    x tick style={color=gray!30},
    tick label style={font=\labelsize},
    ]
\end{axis}
\end{tikzpicture}

\question{Weitere Kommentare zu diesem Abschnitt: {\normalfont(Optional)}}{q:comments-opinions}
\translq{Additional comments on this section: (Optional)}

\surveynote{48 write-in answers.}

\sectionboundary

\noindent
\german{In diesem Abschnitt fragen wir Sie zu Ihrer Meinung zu verschiedenen Sprachtechnologien. \textbf{Dabei gibt es keine richtigen\fslash{}falschen oder erwünschten\fslash{}unerwünschten
Antworten,} sondern wir sind an Ihrer persönlichen Meinung interessiert.}
\translation{In this section we ask you about your opinion on different language technologies. \textbf{There are no right/wrong or desired/undesired answers,} we are simply interested in your personal opinion.}

\definition{\textbf{Übersetzungsprogramme} erstellen eine automatische Übersetzung von Text aus einer Sprache in eine andere Sprache. Beispiele dafür sind DeepL oder Google Translate.}
{\textbf{Machine translation software} automatically translate text from one language into another. Examples are DeepL or Google Translate.}

\question{Stimmen Sie den folgenden Aussagen zu? Es sollte Übersetzungsprogramme geben, ...}{q:mt}
\translq{Do you agree with the following statements? There should be translation software...}

\begin{itemize}
    \item ...die hochdeutsche Texte in meinen Dialekt übersetzen. \translation{...that translates Standard German texts into my dialect.}
    \item ...die Texte aus anderen Sprachen in meinen Dialekt übersetzen. \translation{...that translates texts from other languages into my dialect.}
    \item ...die Texte aus meinem Dialekt ins Hochdeutsche übersetzen. \translation{...that translates texts from my dialect into Standard German.}
    \item ...die Texte aus meinem Dialekt in andere Sprachen übersetzen. \translation{...that translates texts from my dialect into other languages.}
\end{itemize}

\surveynote{Answer options:}
\begin{itemize}
    \item Ja, unbedingt \translation{Yes, absolutely}
    \item Eher ja \translation{Rather yes}
    \item Weder noch \translation{Neither/nor}
    \item Eher nein \translation{Rather no}
    \item Nein, das halte ich nicht für sinnvoll \translation{No, I don't think this is useful}
    \item Das kann ich nicht bewerten \translation{I cannot judge this}
\end{itemize}

\noindent\surveynote{See Figure~\ref{fig:opinions-lt} for answer distributions.}

\question{Welcher Aussage stimmen Sie mehr zu? Wenn ich einen Text in meinen Dialekt übersetzen lasse, ...}{q:mt-variation}
\translq{With which statement do you agree more? When a text is translated into my dialect, ...}

\begin{answerbars}
\entry{1 -- ...soll die Übersetzung}{sprachlich meiner...*}{52}
\onelineentry{2}{61}
\onelineentry{3}{67}
\onelineentry{4}{32}
\entry{5 -- ...ist es mir egal,}{welcher (geschriebenen)...**}{43}
\entry{Ich möchte keinen geschrie-}{benen Dialekt-Output.***}{72}
\end{answerbars}

{\asterisknote
*\german{1 -- ...soll die Übersetzung sprachlich meiner (geschriebenen) Version des Dialekts voll und ganz entsprechen.}
\translation{1~--~...the translation should fully correspond to my own (written) version of the dialect.}\par

**\german{5 -- ...ist es mir egal, welcher (geschriebenen) Form meines Dialekts die Übersetzung sprachlich entspricht.}
\translation{5~-- ...I do not care which (written) version of my dialect the translation corresponds to.}\par

***\translation{I do not want any written dialect output.}\par
}

\filbreak
\question{Weitere Kommentare zu Übersetzungsprogrammen: {\normalfont(Optional)}}{q:comments-mt}
\translq{Additional comments on machine translation software: (Optional)}

\surveynote{41 write-in answers.}

\definition{\textbf{Rechtschreib- und Grammatikkorrekturprogramme} markieren oder korrigieren mögliche Fehler in Texten, zum Beispiel bei der Eingabe in Microsoft Word.}
{\textbf{Spell- and grammar checkers} highlight or fix potential errors in texts, for instance when writing text in Microsoft Word.}

\question{Stimmen Sie der folgenden Aussage zu? Es sollte Rechtschreib- und Grammatikkorrekturprogramme für meinen Dialekt geben.}{q:spellcheckers}
\translq{Do you agree with the following statement? There should be spell- and grammar checkers for my dialect.}

\surveynote{Same answer options as for question~\ref{q:mt}.
See Figure~\ref{fig:opinions-lt} for the answer distribution.}

\question{Weitere Kommentare zu Rechtschreib- und Grammatikkorrekturprogrammen: {\normalfont(Optional)}}{q:comments-spellcheckers}
\translq{Additional comments on spell- and grammar checkers: (Optional)}

\surveynote{51 write-in answers.}

\definition{\textbf{Transkriptionsprogramme} verschriftlichen gesprochene Sprache. Sie finden beispielsweise bei automatisch erzeugten Untertiteln oder bei Diktiergeräten Einsatz.}
{\textbf{Speech-to-text systems} transcribe spoken language. They are for instance used for automatically generating subtitles or in the context of dictation software.}

\question{Stimmen Sie den folgenden Aussagen zu? Es sollte Transkriptionsprogramme geben, ...}{q:stt}
\translq{Do you agree with the following statements? There should be speech-to-text software...}

\begin{itemize}
    \item \german{...die Audioaufnahmen in meinem Dialekt als geschriebenes Hochdeutsch wiedergeben.} \translation{...that transcribes audio recorded in my dialect as written Standard German.}
    \item \german{...die Audioaufnahmen in meinem Dialekt als geschriebenen Dialekt wiedergeben.} \translation{...that transcribes audio recorded in my dialect as written dialect.}
\end{itemize}

\noindent
\surveynote{Same answer options as for question~\ref{q:mt}.
See Figure~\ref{fig:opinions-lt} for answer distributions.}

\question{Weitere Kommentare zu Transkriptionsprogrammen: {\normalfont(Optional)}}{q:opinions-stt}
\translq{Additional comments on speech-to-text software: (Optional)}

\surveynote{33 write-in answers.}

\definition{\textbf{Text-to-Speech-Systeme} funktionieren umgekehrt wie Transkriptionsprogramme: sie
erzeugen gesprochene Versionen von geschriebenem Text. Ein Beispiel dafür sind
Bildschirmleseprogramme.}
{\textbf{Text-to-speech systems} work the other way around as speech-to-text systems: they generate spoken versions of written text. One example are screen readers.}

\question{Stimmen Sie der folgenden Aussage zu? Es sollte Text-to-Speech-Systeme geben, die meinen Dialekt von geschriebener Form in gesprochene Form umwandeln.}{q:tts}
\translq{Do you agree with the following statement? There should be text-to-speech systems that synthesize dialectal audio for text written in my dialect.}

\surveynote{Same answer options as for question~\ref{q:mt}.
See Figure~\ref{fig:opinions-lt} for the answer distribution.}

\question{Weitere Kommentare zu Text-to-Speech-Systemen: {\normalfont(Optional)}}{q:comments-tts}
\translq{Additional comments on text-to-speech systems: (Optional)}

\surveynote{22 write-in answers.}

\definition{\textbf{Sprachassistenten} sind Programme, die geschriebene oder gesprochene Fragen
beantworten bzw. Befehle ausführen, zum Beispiel Siri oder Alexa.\\ Eng verwandt damit sind \textbf{Chatbots}: Programme, die textbasierte Dialoge ermöglichen, bei denen ein Programm Antworten auf Texteingaben von Nutzer*innen erzeugt. Ein Beispiel dafür ist ChatGPT.\\}
{\textbf{Digital assistants} are programs that answer written or spoken questions and carry out commands, like Siri or Alexa. \\
\textbf{Chatbots} are closely related. They are software that enables text-based dialogues, wherein a program generates answers to text input from users. An example is ChatGPT.}

\question{Stimmen Sie den folgenden Aussagen zu?}{q:assistant}
\translq{Do you agree with the following statements?}

\begin{itemize}
    \item \german{Es sollte Sprachassistenten geben, die man mit Fragen\fslash{}Befehlen in meinem Dialekt bedienen kann.} \translation{There should be digital assistants that you can query with questions\fslash{}commands in my dialect.}
    \item \german{Es sollte Sprachassistenten geben, die in meinem Dialekt auf Fragen\fslash{}Befehle antworten.} \translation{There should be digital assistants that use my dialect when replying to questions\fslash{}commands.}
    \item \german{Es sollte Chatbots geben, die auf Eingaben in meinem Dialekt antworten können.} \translation{There should be chatbots that can respond to inputs written in my dialect.}
    \item \german{Es sollte Chatbots geben, deren Antworten in meinem Dialekt verfasst sind.} \translation{There should be chatbots who respond in my dialect.}
\end{itemize}

\noindent
\surveynote{Same answer options as for question~\ref{q:mt}.
See Figure~\ref{fig:opinions-lt} for answer the distributions.}

\question{Welcher Aussage stimmen Sie mehr zu? Wenn ein Sprachassistent oder ein Chatbot Antworten in meinem Dialekt erzeugt, ...}{q:assistant-variation}
\translq{With which statement do you agree more? When a digital assistant or chatbot generates replies in my dialect, ...}

\begin{answerbars}
\entry{1 -- ...sollen die Antworten}{sprachlich meiner...*}{46}
\onelineentry{2}{69}
\onelineentry{3}{61}
\onelineentry{4}{28}
\entry{5 -- ...ist es mir egal, welcher}{(geschriebenen oder...**}{55}
\entry{Ich möchte keinen geschrie-}{benen Dialekt-Output.***}{68}
\end{answerbars}

{\asterisknote
*\german{1 -- ...sollen die Antworten sprachlich meiner (geschriebenen oder gesprochenen) Version des Dialekts voll und ganz entsprechen.}
\translation{1~--~... the replies should fully correspond to my own (written or spoken) version of the dialect.}\par

**\german{5 -- ...ist es mir egal, welcher (geschriebenen oder gesprochenen) Form meines Dialekts die Antworten sprachlich entsprechen.}
\translation{5~-- ...I do not care which (written or spoken) version of my dialect the replies correspond to.}\par

***\translation{I do not want any written dialect output.}\par
}

\question{Weitere Kommentare zu Sprachassistenten oder Chatbots: {\normalfont(Optional)}}{q:comments-assistants}
\translq{Additional comments on digital assistants or chatbots: (Optional)}

\surveynote{25 write-in answers.}

\definition{\textbf{Suchmaschinen} sind Programme, die nach einer Suchanfrage Datenbanken oder das
Internet nach relevanten Ergebnissen durchsuchen, wie zum Beispiel Google.}
{\textbf{Search engines} are programs that search a database or the web based on a search query, like Google.}

\question{Stimmen Sie der folgenden Aussage zu? Es sollte Suchmaschinen geben, bei denen ich meinen Dialekt als Eingabesprache verwenden kann.}{q:search}
\translq{Do you agree with the following statement? There should be search engines that support queries in my dialect.}

\surveynote{Same answer options as for question~\ref{q:mt}.
See Figure~\ref{fig:opinions-lt} for the answer distribution.}

\question{Weitere Kommentare zu Suchmaschinen: {\normalfont(Optional)}}{q:comments-search}
\translq{Additional comments on search engines: (Optional)}

\surveynote{15 write-in answers.}

\sectionboundary

\question{Sind Ihnen bereits Sprachtechnologien bekannt, die Ihren Dialekt unterstützen?}{q:know-existing}
\translq{Are you already aware of any language technologies for your dialect?}

\begin{answerbars}
\entry{Ja und zwar:}{\translation{Yes, namely:}}{46}
\entry{Nein}{\translation{No}}{280}
\end{answerbars}

\question{Stimmen Sie den folgenden Aussagen zu?}{q:lt-opinions}
\translq{Do you agree with the following statements?}

\surveynote{Statements presented in a randomized order:}

\begin{itemize}
    \item 
\german{Sprachtechnologie, die ich für sinnvoll halte, nutze ich auch selbst.}
\translation{If I find language technology useful, I also use it myself.}

    \item 
\german{Eine größere Unterstützung durch Sprachtechnologie würde meinen Dialekt attraktiver für jüngere
Generationen machen.}
\translation{If my dialect were supported more by language technologies, the dialect would be more appealing for younger generations.}
\surveynote{This question is modelled after questions in the surveys by \citet{soria-etal-2018-dldp} and \citet[p.~229]{millour2020myriadisation}, asking about the hypothesized impact of a language's increased use online on the appeal for younger people.}

    \item 
\german{Wenn ich Sprachtechnologie für meinen Dialekt hätte, würde ich ihn häufiger verwenden.}
\translation{If I had language technology for my dialect, I would use my dialect more often.}
\end{itemize}

\noindent\surveynote{See question~\ref{q:opinions-dialect} for the answer options. Answer distributions (in \%):}
\pgfplotscreateplotcyclelist{customcols}{darkred, darkred!40, gray!40, gray!20, mydarkblue!40, mydarkblue}

\pgfplotstableread[col sep=comma]{fig/opinions_end.csv}\opiniondata

\setlength{\plotheight}{4cm}
\setlength{\plotwidth}{0.95\columnwidth}

\begin{tikzpicture}[font=\sffamily]
\begin{axis}[
    width=\plotwidth, height=\plotheight,
    xbar stacked,
    xmin=0, xmax=100,
    axis line style={draw=none},
    ytick=data,
    yticklabels from table={\opiniondata}{LT},
    yticklabel style={align=right},
    enlarge y limits={abs=4mm},
    ytick style={draw=none},
    tick label style={font=\labelsize},
    xtick={0, 20, 40, 60, 80, 100},
    xticklabels={0, 20, 40, \gr{60}, \gr{80}, \gr{100\,\%}},
    grid=major, tick align=inside,
    major grid style={draw=gray!30},
    x tick style={color=gray!30},
    legend style={
        legend columns=2,
        transpose legend,
        at={(xticklabel cs:0.4)},
        anchor=north,
        draw=none,
        /tikz/every even column/.append style={column sep=0.25cm},
        font=\labelsize,
        },
    legend cell align={left},
    cycle list name=customcols,
    every axis plot/.style={fill},
    nodes near coords style={font=\labelsize, /pgf/number format/assume math mode, /pgf/number format/.cd, fixed, precision=0},
    nodes near coords, 
    ]
\addplot+[text=black!15] table[x=No, meta=LT,y expr=\thisrow{YPos}] {\opiniondata};
\addplot+[text=black!5] table[x=RatherNo, meta=LT,y expr=\thisrow{YPos}] {\opiniondata};
\addplot+[text=black!70] table[x=CannotJudge, meta=LT,y expr=\thisrow{YPos}] {\opiniondata};
\addplot+[text=black!50] table[x=NeitherNor, meta=LT,y expr=\thisrow{YPos}] {\opiniondata};
\addplot+[text=black!5] table[x=RatherYes, meta=LT,y expr=\thisrow{YPos}] {\opiniondata};
\addplot+[text=black!15] table[x=Yes, meta=LT,y expr=\thisrow{YPos}] {\opiniondata};
\legend{Disagree, Dis.\ somewhat, N/A, Neither/nor, A.\ somewhat, Agree}
\end{axis}
\begin{axis}[
    width=\plotwidth, height=\plotheight,
    axis line style={draw=none},
    xmin=0, xmax=100,
    xtick={0, 20, 40, 60, 80, 100},
    xticklabels={, \gr{100}, \gr{80}, \gr{60}, 40, 20, 0\,\%},
    axis x line*=top,
    ytick=none,
    yticklabels={},
    y tick style={opacity=0},
    x tick style={color=gray!30},
    tick label style={font=\labelsize},
    ]
\end{axis}
\end{tikzpicture}

\question{Weitere Kommentare zum Thema Sprachtechnologie oder allgemein zu dieser
Umfrage: {\normalfont(Optional)}}{q:comments-rest}
\translq{Additional comments on language technology or generally regarding this survey: (Optional)}

\surveynote{29 write-in answers.}

\question{Wie haben Sie von dieser Studie erfahren?}{q:source}
\translq{How did you find out about this study?}

\begin{answerbars}
\entry{Durch Forschende am}{Centrum für ...*}{41}
\entry{Durch}{Mundartpflegevereine**}{20}
\entry{Soziale Medien, Foren}{\translation{Social media, forums}}{46}
\entry{E-Mail-Verteiler}{\translation{Mailing lists}}{119}
\entry{(Anderweitig) durch}{Bekannte***}{79}
\entry{Auf eine andere Art:}{\translation{Otherwise:}}{20}
\end{answerbars}

{\asterisknote
*\german{Durch Forschende am Centrum für Informations- und Sprachverarbeitung (LMU)}
\translation{Via researchers at the Center for Information and Language Processing (LMU)}\par

**\translation{Via dialect preservation societies}\par

***\translation{(Otherwise) via acquaintances}\par
}

\sectionboundary

\noindent
\textbf{Vielen Dank für Ihre Teilnahme!}
Wir möchten uns ganz herzlich für Ihre Mithilfe bedanken.
Ihre Antworten wurden gespeichert, Sie können das Browser-Fenster nun schließen.
\translation{\textbf{Thank you for participating!} We would like to thank you very much for your help. Your answers have been saved; you can close the browser window now.}

\newcommand{\ques}[1]{{(\ref{#1})}}

\newcommand{\dialskills}{Dialect skills \ques{q:skills}}
\newcommand{\aoa}{Age of acquisition \ques{q:aoa}}
\newcommand{\age}{Age \ques{q:age}}
\newcommand{\activism}{Activism \ques{q:activism}}
\newcommand{\trad}{Traditionality \ques{q:traditionality}}
\newcommand{\freq}{Frequency \ques{q:freq}}
\newcommand{\writing}{Writing \ques{q:writing}}
\newcommand{\nrwriting}{\# Writing contexts  \ques{q:write-what}}
\newcommand{\aspect}{Any aspect \ques{q:opinions-dialect}}
\newcommand{\ortho}{Orthography \ques{q:opinions-dialect}}
\newcommand{\spokenonly}{Spoken only \ques{q:opinions-dialect}}
\newcommand{\diversity}{Diversity \ques{q:opinions-dialect}}
\newcommand{\reading}{Reading is hard \ques{q:opinions-dialect}}
\newcommand{\mtdeudial}{MT deu$\rightarrow$dial \ques{q:mt}}
\newcommand{\mtothdial}{MT oth$\rightarrow$dial \ques{q:mt}}
\newcommand{\mtdialdeu}{MT dial$\rightarrow$deu \ques{q:mt}}
\newcommand{\mtdialoth}{MT dial$\rightarrow$oth \ques{q:mt}}
\newcommand{\mtvar}{MT variation \ques{q:mt-variation}}
\newcommand{\spell}{Spellcheckers \ques{q:spellcheckers}}
\newcommand{\sttdeu}{STT deu \ques{q:stt}}
\newcommand{\sttdial}{STT dial \ques{q:stt}}
\newcommand{\tts}{TTS \ques{q:tts}}
\newcommand{\assistantin}{Assistant in \ques{q:assistant}}
\newcommand{\assistantout}{Assistant out \ques{q:assistant}}
\newcommand{\chatbotin}{Chatbot in \ques{q:assistant}}
\newcommand{\chatbotout}{Chatbot out \ques{q:assistant}}
\newcommand{\assistantvar}{Assistant var. \ques{q:assistant-variation}}
\newcommand{\search}{Search engines \ques{q:search}}
\newcommand{\knows}{Know existing \ques{q:know-existing}}
\newcommand{\appeal}{Appeal \ques{q:lt-opinions}}
\newcommand{\useful}{Useful=use \ques{q:lt-opinions}}
\newcommand{\often}{More often \ques{q:lt-opinions}}

\filbreak
\begin{table*}[t]
\centering
\begin{tabular}{@{}rl@{\hspace{-2pt}}rclr@{}}
\toprule
 & \multicolumn{2}{c}{\textbf{All}} && \multicolumn{2}{c}{\textbf{Non-activists only}} \\ \cmidrule(l){2-3} \cmidrule(l){5-6}  
\textbf{Rank} & \textbf{LTs} & \textbf{Mean} && \textbf{LTs} & \textbf{Mean} \\ \midrule
1 & \assistantin & 3.75 && \assistantin & 3.80 \\
2 & \sttdeu & 3.46 && \sttdeu & 3.48 \\
3 & \sttdial & 3.38\raisebox{4pt}{\tikzmark{threeL}}&  & \raisebox{4pt}{\tikzmark{threeR}}\chatbotin & 3.25 \\
4 & \chatbotin & 3.29\raisebox{4pt}{\tikzmark{fourL}}&  & \raisebox{4pt}{\tikzmark{fourR}}\sttdial & 3.24 \\
5 & \mtdialdeu & 3.17\raisebox{4pt}{\tikzmark{fiveL}} && \raisebox{4pt}{\tikzmark{fiveR}}\assistantout & 3.01 \\
6 & \assistantout & 3.14\raisebox{4pt}{\tikzmark{sixL}} && \raisebox{4pt}{\tikzmark{sixR}}\mtdialdeu & 3.00 \\
7 & \tts & 3.13 && \tts & 2.99 \\
8 & \search & 2.94 && \search & 2.69 \\
9 & \chatbotout & 2.76 && \chatbotout & 2.59 \\
10 & \mtdialoth & 2.73 && \mtdialoth & 2.59 \\
11 & \mtdeudial & 2.71 && \mtdeudial & 2.53 \\
12 & \mtothdial & 2.39 && \mtothdial & 2.17 \\
13 & \spell & 2.38 && \spell & 2.08 \\ \bottomrule
\end{tabular}
\begin{tikzpicture}[overlay, remember picture, every path/.style={-, shorten >=4pt, shorten <=4pt}]
\draw (pic cs:threeL) -- (pic cs:fourR);
\draw (pic cs:threeR) -- (pic cs:fourL);
\draw (pic cs:fiveL) -- (pic cs:sixR);
\draw (pic cs:fiveR) -- (pic cs:sixL);
\end{tikzpicture}
\caption{\textbf{Language technologies sorted by mean score} given by all respondents and non-activists only (participants who did not indicate involvement in language preservation,~\S\ref{sec:results-groups}). `Mean' refers to the mean Likert score (see text). Numbers behind the LT names refer to questions in~\S\ref{sec:full-questionnaire}.}
\label{tab:ranking}
\end{table*}

\section{Correlation Scores}
\label{sec:correlations}

Figure~\ref{fig:correlations} shows the Spearman's rank correlation coefficients (\myrho) between the variables investigated in the questionnaire, with \myrho~values ranging from --0.50 to +0.77.

For the correlation analysis and the subgroup comparisons (Appendix~\S\ref{sec:subgroup-comparisons}), the variable values are mapped so that higher values correspond to higher agreement with the statements in questions \ref{q:opinions-dialect}, \ref{q:mt}, \ref{q:spellcheckers}, \ref{q:stt}, \ref{q:tts}, \ref{q:assistant}, \ref{q:search} and~\ref{q:lt-opinions}, and to higher dialect competence (question~\ref{q:skills}) and usage frequency~(\ref{q:freq}), higher age~(\ref{q:age}) and age of dialect acquisition~(\ref{q:aoa}), more traditional dialects (\ref{q:traditionality}),\footnote{Note that this the inverse of how the question is originally phrased.} and greater openness towards variation in the output of MT~(\ref{q:mt-variation}) and digital assistants\,/\,\allowbreak{}chatbots~(\ref{q:assistant-variation}).
The variable \textit{\#~writing contexts} encodes the number of answer options selected in question~\ref{q:write-what}.
The variables \textit{writing}~(\ref{q:writing}) and \textit{activism}~(\ref{q:activism}) are binary such that 0~encodes the `no' options and 1~stands for the `yes' options.

The beginning of the first row in the figure can thus be read as follows:
Dialect competence self-ratings are
\begin{itemize}
    \item negatively correlated with the age of acquisition (i.e., respondents whose dialect is their first language generally give higher competence ratings), 
    \item slightly positively correlated with language activism (i.e., fluent dialect speakers are slightly more likely to be engaged in dialect preservation activities, and vice versa),
    \item positively correlated with traditionality (i.e., competent dialect speakers tend to rate their dialect as more distinct from Standard German, and vice versa),
\end{itemize}
\noindent
and so on.

\section{LT Ranking}
\label{sec:ranking}
Table~\ref{tab:ranking} shows the order of preferred LTs.
This ranking is based on the mean scores when coding the answers as follows:
1\,=\,useless,
2\,=\,rather useless,
3\,=\,neither/not,
4\,=\,rather useful,
5\,=\,useful.
Non-answers (`cannot judge') are excluded.

If we remove the participants who indicated active engagement in dialect preservation (see \S\ref{sec:results-groups} and question~\ref{q:activism}), the ranking only changes very slightly: chatbots with dialectal input and STT with dialectal output trade places (although they have nearly identical mean scores), and we observe the same for virtual assistants with dialectal output and machine translation from the dialect into Standard German.

\section{Subgroup Comparisons}
\label{sec:subgroup-comparisons}

\newcommand{\insignificant}[1]{\textcolor{black!30}{#1}}
\newcommand{\colsp}{\hspace{30pt}}
\newcommand{\sigdist}{\hspace{2pt}}
\newcommand{\notapplic}{\rule[0.8ex]{1.4em}{0.7pt}}

\newcommand{\sigi}{{\scriptsize *}}
\newcommand{\sigii}{{\scriptsize **}}
\newcommand{\sigiii}{{\scriptsize ***}}

\newcommand{\chisq}{$\chi^{2}$}
\newcommand{\tstat}{\textit{t-}stat}

Tables~\ref{tab:subgroups-activists} and~\ref{tab:subgroups-region} show how the responses by different subgroups of respondents differ for each variable. We provide each subgroup's mean answers (using the same numeric coding as in the previous two appendix sections), as well as \textit{t-}test statistics (taking into account the scalar nature of the answer options) and \chisq{} test results.

In addition to the analyses in \S\ref{sec:results-groups}, we provide two more subgroup comparisons, albeit with small effect sizes:

\paragraph{Traditionality}
We compare the responses of speakers who rate their dialect as traditional and distinct from Standard German (the first two answer options for question~\ref{q:traditionality}) to those who indicated speaking a variety more akin to a regiolect (the last two options).
While these subgroups differ in their responses to the dialect-related questions, few of the differences regarding language technologies are statistically significant (Table~\ref{tab:subgroups-activists}).

\paragraph{Age}
Figure~\ref{fig:correlations} shows that the variable \textit{age} correlates with few other variables.
With respect to the LTs, young participants tend to be somewhat more positive towards three of the overall most popular LTs: STT with Standard German output, and virtual assistants and chatbots with dialectal input.

\clearpage
\newcommand{\ticklabels}{{\dialskills, \aoa, \age, \activism, \trad, \freq, \writing, \nrwriting, \aspect, \ortho, \spokenonly, \diversity, \reading, \mtdeudial, \mtothdial, \mtdialdeu, \mtdialoth, \mtvar, \spell, \sttdeu, \sttdial, \tts, \assistantin, \assistantout, \chatbotin, \chatbotout, \assistantvar, \search, \knows, \appeal,\useful, \often}}

\begin{figure*}
\begin{adjustbox}{max width=\textwidth}
\begin{tikzpicture}
\begin{axis}[
    width=\textwidth, height=\textwidth,
    scatter,
    colormap/RdBu,
    point meta max=0.77, %
    point meta min=-0.77, %
    y dir=reverse,
    axis line style={draw=none}, %
    grid=major,
    tickwidth=0pt,
    xtick=data,
    ytick=data,
    xticklabels/.expanded=\ticklabels,
    yticklabels/.expanded=\ticklabels,
    xticklabel style={rotate=90},
    enlargelimits={abs=0.5},
    visualization depends on=\thisrow{size}\as\psize,
    scatter/@pre marker code/.append code={
      \scope[mark size=\psize, fill=mapped color]
    },
    scatter/@post marker code/.append code={\endscope},
]
\addplot +[
    point meta=explicit, %
    only marks, %
    ] table [
    meta=correlation
] {fig/correlations.tsv};
\end{axis}
\end{tikzpicture}
\end{adjustbox}

\hfill
\begin{tikzpicture}
\begin{axis}[
    scatter,
    title={Correlation (\myrho)},
    title style={xshift=1.5em, yshift=-0.5em, font=\footnotesize},
    colormap/RdBu-11,
    point meta max=0.77, %
    point meta min=-0.77, %
    width=0.15\textwidth,
    height=0.27\textwidth,
    axis line style={draw=none},
    grid=none,
    tickwidth=0pt,
    ytick=data,
    tick label style={font=\footnotesize},
    yticklabels={\enspace0.75, \enspace0.50, \enspace0.25, \enspace0.00, --0.25, --0.50},
    xticklabels={},
    axis y line*=right,
    scatter/@pre marker code/.append code={
      \scope[mark size=4, fill=mapped color]
    },
    scatter/@post marker code/.append code={\endscope},
]
\addplot +[point meta=explicit, only marks,] table [meta=correlation] {
x   y   correlation
0   0   0.75
0   -1   0.5
0   -2   0.25
0   -3   0
0   -4   -0.25
0   -5   -0.5
};
\end{axis}
\end{tikzpicture}
\begin{tikzpicture}
\begin{axis}[
    scatter,
    title={\textit{p}-value},
    title style={xshift=2em, yshift=-0.5em, font=\footnotesize},
    colormap/RdBu,
    point meta max=0.77, %
    point meta min=-0.77, %
    width=0.15\textwidth,
    height=0.27\textwidth,
    axis line style={draw=none},
    grid=none,
    tickwidth=0pt,
    ytick=data,
    tick label style={font=\footnotesize},
    yticklabels={<0.001, <0.01, <0.05, \mygeq0.05, , },
    xticklabels={},
    axis y line*=right,
    visualization depends on=\thisrow{size}\as\psize,
    scatter/@pre marker code/.append code={
      \scope[mark size=\psize, fill=black]
    },
    scatter/@post marker code/.append code={\endscope},
]
\addplot +[only marks,] table {
x   y   size
0   0   4.95
0   -1   3.6
0   -2   2.25
0   -3   0
0 -4 0
0 -5 0
};
\end{axis}
\end{tikzpicture}
\vspace{-8.7em}

\captionsetup{margin={0pt, 14.5em}}
\caption{\textbf{Spearman's \bm{\myrho} between variables.} 
Blue dots show positively correlated variables (max.: +0.77), red dots negatively correlated ones (min.: -0.50). 
We only include correlations with \textit{p-}values under 0.05.
The larger the dot, the smaller the \textit{p-}value.
The numbers behind the variables refer to the questions in Appendix~\S\ref{sec:full-questionnaire}.
For further explanations of how the variables are coded, see Appendix~\S\ref{sec:correlations}.}
\label{fig:correlations}
\end{figure*}

\begin{table*}
\centering
\renewcommand{\arraystretch}{1.15}
\begin{tabular}{l@{\hspace{15pt}}r@{\sigdist}lr@{\sigdist}lrr@{\colsp}r@{\sigdist}lr@{\sigdist}lrr}
\toprule
\textbf{Variable} & \multicolumn{6}{l}{\textbf{Activists vs. non-activists}} & \multicolumn{6}{l}{\textbf{Most vs. least trad. dialects}} \\
 & \multicolumn{2}{c}{\tstat} & \multicolumn{2}{c}{\chisq}  & {$\mu_{\text{Act}}$} & {$\mu_{\text{Non}}$} & \multicolumn{2}{c}{\tstat} & \multicolumn{2}{c}{\chisq}  & {$\mu_{\text{Most}}$} & {$\mu_{\text{Least}}$} \\
 \midrule
{\dialskills} & 3.0 & {\sigii} & 9.0 & {\sigi} & 3.7 & 3.5 & 11.3 & {\sigiii} & 83.2 & {\sigiii} & 3.8 & 2.9 \\
{\aoa} & {\color[HTML]{B7B7B7} 1.3} &  & {\color[HTML]{B7B7B7} 5.4} &  & 1.4 & 1.3 & -4.4 & {\sigiii} & 29.9 & {\sigiii} & 1.2 & 1.5 \\
{\age} & 6.7 & {\sigiii} & 43.5 & {\sigiii} & 4.7 & 3.4 & {\color[HTML]{B7B7B7} 0.7} &  & {\color[HTML]{B7B7B7} 3.9} &  & 3.9 & 3.7 \\
{\activism} & {\notapplic} &  & {\notapplic} &  & 1.0 & 0.0 & 5.7 & {\sigiii} & 26.9 & {\sigiii} & 0.5 & 0.1 \\
{\trad} & 5.6 & {\sigiii} & 31.6 & {\sigiii} & 3.9 & 3.2 & {\notapplic} &  & {\notapplic} &  & 4.3 & 1.6 \\
{\freq} & 2.5 & {\sigi} & 14.0 & {\sigi} & 5.1 & 4.6 & 5.0 & {\sigiii} & 25.9 & {\sigiii} & 5.1 & 4.0 \\
{\writing} & 3.9 & {\sigiii} & 13.9 & {\sigiii} & 0.8 & 0.6 & 5.0 & {\sigiii} & 21.3 & {\sigiii} & 0.8 & 0.4 \\
{\nrwriting} & 4.9 & {\sigiii} & 28.7 & {\sigiii} & 3.3 & 2.2 & 3.4 & {\sigiii} & 16.2 & {\sigi} & 3.0 & 1.8 \\
{\aspect} & 5.6 & {\sigiii} & 36.3 & {\sigiii} & 3.7 & 2.8 & 6.2 & {\sigiii} & 36.4 & {\sigiii} & 3.5 & 2.2 \\
{\ortho} & 6.0 & {\sigiii} & 37.9 & {\sigiii} & 2.8 & 2.0 & 2.8 & {\sigii} & 12.6 & {\sigi} & 2.5 & 1.9 \\
{\spokenonly} & -4.2 & {\sigiii} & 34.9 & {\sigiii} & 2.9 & 3.5 & -2.5 & {\sigi} & {\color[HTML]{B7B7B7} 8.7} &  & 3.1 & 3.7 \\
{\diversity} & 3.0 & {\sigii} & 12.5 & {\sigi} & 4.3 & 3.9 & 3.0 & {\sigii} & 12.5 & {\sigi} & 4.2 & 3.7 \\
{\reading} & {\color[HTML]{B7B7B7} -1.6} &  & {\color[HTML]{B7B7B7} 5.5} &  & 2.2 & 2.4 & -2.3 & {\sigi} & 10.4 & {\sigi} & 2.3 & 2.7 \\
{\mtdeudial} & 3.0 & {\sigii} & 14.7 & {\sigii} & 3.0 & 2.5 & {\color[HTML]{B7B7B7} 0.9} &  & {\color[HTML]{B7B7B7} 2.0} &  & 2.7 & 2.5 \\
{\mtothdial} & 4.1 & {\sigiii} & 21.8 & {\sigiii} & 2.8 & 2.2 & {\color[HTML]{B7B7B7} 1.6} &  & 12.3 & {\sigi} & 2.5 & 2.2 \\
{\mtdialdeu} & 3.1 & {\sigii} & 13.1 & {\sigi} & 3.5 & 3.0 & {\color[HTML]{B7B7B7} 1.5} &  & {\color[HTML]{B7B7B7} 7.9} &  & 3.2 & 2.9 \\
{\mtdialoth} & 2.4 & {\sigi} & {\color[HTML]{B7B7B7} 7.0} &  & 3.0 & 2.6 & 2.0 & {\sigi} & 20.0 & {\sigiii} & 2.9 & 2.4 \\
{\mtvar} & -3.3 & {\sigii} & 11.7 & {\sigi} & 2.5 & 3.0 & -4.2 & {\sigiii} & 19.1 & {\sigiii} & 2.6 & 3.6 \\
{\spell} & 5.4 & {\sigiii} & 34.2 & {\sigiii} & 2.9 & 2.1 & 2.0 & {\sigi} & 9.8 & {\sigi} & 2.4 & 2.0 \\
{\sttdeu} & {\color[HTML]{B7B7B7} -0.3} &  & {\color[HTML]{B7B7B7} 3.1} &  & 3.4 & 3.5 & {\color[HTML]{B7B7B7} 1.0} &  & {\color[HTML]{B7B7B7} 8.9} &  & 3.5 & 3.3 \\
{\sttdial} & 2.4 & {\sigi} & {\color[HTML]{B7B7B7} 8.7} &  & 3.6 & 3.2 & 2.6 & {\sigii} & 10.1 & {\sigi} & 3.5 & 2.9 \\
{\tts} & 2.5 & {\sigi} & 11.6 & {\sigi} & 3.4 & 3.0 & {\color[HTML]{B7B7B7} 1.0} &  & {\color[HTML]{B7B7B7} 6.9} &  & 3.1 & 2.9 \\
{\assistantin} & {\color[HTML]{B7B7B7} -0.8} &  & {\color[HTML]{B7B7B7} 3.3} &  & 3.7 & 3.8 & {\color[HTML]{B7B7B7} 1.0} &  & {\color[HTML]{B7B7B7} 1.6} &  & 3.8 & 3.6 \\
{\assistantout} & 2.3 & {\sigi} & {\color[HTML]{B7B7B7} 5.6} &  & 3.4 & 3.0 & {\color[HTML]{B7B7B7} -0.1} &  & {\color[HTML]{B7B7B7} 1.7} &  & 3.1 & 3.1 \\
{\chatbotin} & {\color[HTML]{B7B7B7} 0.7} &  & {\color[HTML]{B7B7B7} 1.3} &  & 3.4 & 3.2 & {\color[HTML]{B7B7B7} 0.2} &  & {\color[HTML]{B7B7B7} 3.7} &  & 3.3 & 3.3 \\
{\chatbotout} & 3.1 & {\sigii} & 12.1 & {\sigi} & 3.1 & 2.6 & {\color[HTML]{B7B7B7} 0.6} &  & {\color[HTML]{B7B7B7} 2.0} &  & 2.8 & 2.7 \\
{\assistantvar} & -3.5 & {\sigiii} & 13.7 & {\sigii} & 2.5 & 3.1 & -4.9 & {\sigiii} & 22.9 & {\sigiii} & 2.7 & 3.7 \\
{\search} & 4.2 & {\sigiii} & 20.8 & {\sigiii} & 3.4 & 2.7 & {\color[HTML]{B7B7B7} 0.9} &  & {\color[HTML]{B7B7B7} 4.2} &  & 2.9 & 2.7 \\
{\knows} & {\color[HTML]{B7B7B7} 0.6} &  & {\color[HTML]{B7B7B7} 0.2} &  & 0.2 & 0.1 & 2.0 & {\sigi} & {\color[HTML]{B7B7B7} 3.1} &  & 0.2 & 0.1 \\
{\appeal} & 5.4 & {\sigiii} & 32.1 & {\sigiii} & 4.0 & 3.2 & {\color[HTML]{B7B7B7} 1.5} &  & {\color[HTML]{B7B7B7} 8.0} &  & 3.6 & 3.3 \\
{\useful} & {\color[HTML]{B7B7B7} 0.5} &  & {\color[HTML]{B7B7B7} 5.0} &  & 3.7 & 3.6 & {\color[HTML]{B7B7B7} 1.3} &  & {\color[HTML]{B7B7B7} 3.4} &  & 3.7 & 3.5 \\
{\often} & 4.0 & {\sigiii} & 16.5 & {\sigii} & 3.1 & 2.5 & {\color[HTML]{B7B7B7} 0.1} &  & {\color[HTML]{B7B7B7} 1.0} &  & 2.6 & 2.6\\
\bottomrule
\end{tabular}
\caption{\textbf{Differences between respondent subgroups.}
We show the results of \textit{t-}tests and {\chisq} tests between pairs of respondent groups: those who indicated involvement in dialect preservation efforts (`activists', question~\ref{q:activism}) vs.\ those who did not, and respondents who rate their dialect as one of the two most vs.\ two least traditional options (question~\ref{q:traditionality}).
Positive \textit{t-}statistics indicate that the first group's values for the variable are higher than the second one's, and vice versa for negative values.
\insignificant{Grey entries} denote results with \textit{p-}values $\geq0.05$; asterisks represent smaller \textit{p-}values: {\small*}\,$<0.05$, {\small**}\,$<0.01$, {\small***}\,$<0.001$.
The columns with $\mu$ present the mean Likert scores of the subgroups' responses (e.g., $\mu_{\text{Act}}$ contains the activists' mean answers).
The numbers behind the variables refer to the questions in Appendix~\S\ref{sec:full-questionnaire}.
For information on the variables on how the variables are encoded as numbers, see Appendix~\S\ref{sec:correlations}.}
\label{tab:subgroups-activists}
\end{table*}

\begin{table*}
\centering
\renewcommand{\arraystretch}{1.15}
\begin{tabular}{l@{\hspace{15pt}}r@{\sigdist}lr@{\sigdist}l@{\colsp}r@{\sigdist}lr@{\sigdist}l@{\colsp}rrr}
\toprule
 & \multicolumn{4}{l}{\textbf{NDS vs. (rest of) D/AT}} & \multicolumn{7}{l}{\textbf{CH vs. (non-NDS) D/AT}}  \\
 & \multicolumn{2}{c}{\tstat} & {\chisq} & & \multicolumn{2}{c}{\tstat}  & {\chisq}  && {$\mu_{\text{NDS}}$} & {$\mu_{\text{D/AT}}$} & {$\mu_{\text{CH}}$} \\
 \midrule
{\dialskills} & {\color[HTML]{B7B7B7} -1.6} & {\color[HTML]{000000} } & {\color[HTML]{B7B7B7} 5.5} & {\color[HTML]{000000} } & {\color[HTML]{000000} 4.7} & {\color[HTML]{000000} {\sigiii}} & {\color[HTML]{000000} 23.4} & {\color[HTML]{000000} {\sigiii}} & {\color[HTML]{000000} 3.4} & {\color[HTML]{000000} 3.5} & 4.0 \\
{\aoa} & {\color[HTML]{000000} 5.2} & {\color[HTML]{000000} {\sigiii}} & {\color[HTML]{000000} 25.2} & {\color[HTML]{000000} {\sigiii}} & {\color[HTML]{000000} -3.2} & {\color[HTML]{000000} {\sigii}} & {\color[HTML]{000000} 10.0} & {\color[HTML]{000000} {\sigii}} & {\color[HTML]{000000} 1.7} & {\color[HTML]{000000} 1.3} & 1.1 \\
{\age} & {\color[HTML]{000000} 6.3} & {\color[HTML]{000000} {\sigiii}} & {\color[HTML]{000000} 38.2} & {\color[HTML]{000000} {\sigiii}} & {\color[HTML]{B7B7B7} -1.5} & {\color[HTML]{000000} } & {\color[HTML]{000000} 15.1} & {\color[HTML]{000000} {\sigi}} & {\color[HTML]{000000} 5.3} & {\color[HTML]{000000} 3.6} & 3.2 \\
{\activism} & {\color[HTML]{000000} 5.3} & {\color[HTML]{000000} {\sigiii}} & {\color[HTML]{000000} 23.9} & {\color[HTML]{000000} {\sigiii}} & {\color[HTML]{B7B7B7} -1.8} & {\color[HTML]{000000} } & {\color[HTML]{B7B7B7} 2.5} & {\color[HTML]{000000} } & {\color[HTML]{000000} 0.7} & {\color[HTML]{000000} 0.3} & 0.2 \\
{\trad} & {\color[HTML]{000000} 2.8} & {\color[HTML]{000000} {\sigii}} & {\color[HTML]{000000} 9.9} & {\color[HTML]{000000} {\sigi}} & {\color[HTML]{000000} 3.3} & {\color[HTML]{000000} {\sigii}} & {\color[HTML]{000000} 16.6} & {\color[HTML]{000000} {\sigii}} & {\color[HTML]{000000} 3.7} & {\color[HTML]{000000} 3.3} & 3.8 \\
{\freq} & {\color[HTML]{B7B7B7} -1.1} & {\color[HTML]{000000} } & {\color[HTML]{B7B7B7} 6.4} & {\color[HTML]{000000} } & {\color[HTML]{000000} 4.6} & {\color[HTML]{000000} {\sigiii}} & {\color[HTML]{000000} 23.0} & {\color[HTML]{000000} {\sigiii}} & {\color[HTML]{000000} 4.4} & {\color[HTML]{000000} 4.7} & 5.7 \\
{\writing} & {\color[HTML]{B7B7B7} 1.2} & {\color[HTML]{000000} } & {\color[HTML]{B7B7B7} 1.1} & {\color[HTML]{000000} } & {\color[HTML]{000000} 3.8} & {\color[HTML]{000000} {\sigiii}} & {\color[HTML]{000000} 12.5} & {\color[HTML]{000000} {\sigiii}} & {\color[HTML]{000000} 0.7} & {\color[HTML]{000000} 0.6} & 0.9 \\
{\nrwriting} & {\color[HTML]{000000} 6.5} & {\color[HTML]{000000} {\sigiii}} & {\color[HTML]{000000} 45.9} & {\color[HTML]{000000} {\sigiii}} & {\color[HTML]{000000} 4.5} & {\color[HTML]{000000} {\sigiii}} & {\color[HTML]{000000} 25.5} & {\color[HTML]{000000} {\sigiii}} & {\color[HTML]{000000} 4.0} & {\color[HTML]{000000} 2.1} & 3.2 \\
{\aspect} & {\color[HTML]{000000} 5.3} & {\color[HTML]{000000} {\sigiii}} & {\color[HTML]{000000} 27.6} & {\color[HTML]{000000} {\sigiii}} & {\color[HTML]{000000} 3.9} & {\color[HTML]{000000} {\sigiii}} & {\color[HTML]{000000} 27.3} & {\color[HTML]{000000} {\sigiii}} & {\color[HTML]{000000} 3.9} & {\color[HTML]{000000} 2.8} & 3.7 \\
{\ortho} & {\color[HTML]{000000} 8.4} & {\color[HTML]{000000} {\sigiii}} & {\color[HTML]{000000} 63.6} & {\color[HTML]{000000} {\sigiii}} & {\color[HTML]{000000} -2.0} & {\color[HTML]{000000} {\sigi}} & {\color[HTML]{B7B7B7} 4.8} & {\color[HTML]{000000} } & {\color[HTML]{000000} 3.6} & {\color[HTML]{000000} 2.0} & {\color[HTML]{000000} 1.7} \\
{\spokenonly} & {\color[HTML]{000000} -5.6} & {\color[HTML]{000000} {\sigiii}} & {\color[HTML]{000000} 30.6} & {\color[HTML]{000000} {\sigiii}} & {\color[HTML]{000000} -3.0} & {\color[HTML]{000000} {\sigii}} & {\color[HTML]{000000} 19.7} & {\color[HTML]{000000} {\sigiii}} & {\color[HTML]{000000} 2.5} & {\color[HTML]{000000} 3.6} & {\color[HTML]{000000} 3.0} \\
{\diversity} & {\color[HTML]{B7B7B7} 1.0} & {\color[HTML]{000000} } & {\color[HTML]{B7B7B7} 3.1} & {\color[HTML]{000000} } & {\color[HTML]{000000} 2.7} & {\color[HTML]{000000} {\sigii}} & {\color[HTML]{B7B7B7} 7.3} & {\color[HTML]{000000} } & {\color[HTML]{000000} 4.2} & {\color[HTML]{000000} 4.0} & {\color[HTML]{000000} 4.5} \\
{\reading} & {\color[HTML]{000000} -4.1} & {\color[HTML]{000000} {\sigiii}} & {\color[HTML]{000000} 18.3} & {\color[HTML]{000000} {\sigii}} & {\color[HTML]{B7B7B7} -1.8} & {\color[HTML]{000000} } & {\color[HTML]{B7B7B7} 3.2} & {\color[HTML]{000000} } & {\color[HTML]{000000} 1.8} & {\color[HTML]{000000} 2.6} & {\color[HTML]{000000} 2.2} \\
{\mtdeudial} & {\color[HTML]{000000} 4.6} & {\color[HTML]{000000} {\sigiii}} & {\color[HTML]{000000} 21.3} & {\color[HTML]{000000} {\sigiii}} & {\color[HTML]{B7B7B7} -0.5} & {\color[HTML]{000000} } & {\color[HTML]{B7B7B7} 1.6} & {\color[HTML]{000000} } & {\color[HTML]{000000} 3.5} & {\color[HTML]{000000} 2.5} & {\color[HTML]{000000} 2.4} \\
{\mtothdial} & {\color[HTML]{000000} 5.0} & {\color[HTML]{000000} {\sigiii}} & {\color[HTML]{000000} 27.8} & {\color[HTML]{000000} {\sigiii}} & {\color[HTML]{B7B7B7} -0.2} & {\color[HTML]{000000} } & {\color[HTML]{B7B7B7} 5.0} & {\color[HTML]{000000} } & {\color[HTML]{000000} 3.2} & {\color[HTML]{000000} 2.2} & {\color[HTML]{000000} 2.2} \\
{\mtdialdeu} & {\color[HTML]{000000} 2.6} & {\color[HTML]{000000} {\sigii}} & {\color[HTML]{000000} 9.9} & {\color[HTML]{000000} {\sigi}} & {\color[HTML]{B7B7B7} 1.0} & {\color[HTML]{000000} } & {\color[HTML]{B7B7B7} 4.3} & {\color[HTML]{000000} } & {\color[HTML]{000000} 3.6} & {\color[HTML]{000000} 3.0} & {\color[HTML]{000000} 3.3} \\
{\mtdialoth} & {\color[HTML]{000000} 3.3} & {\color[HTML]{000000} {\sigii}} & {\color[HTML]{000000} 13.6} & {\color[HTML]{000000} {\sigii}} & {\color[HTML]{B7B7B7} 1.3} & {\color[HTML]{000000} } & {\color[HTML]{B7B7B7} 4.4} & {\color[HTML]{000000} } & {\color[HTML]{000000} 3.2} & {\color[HTML]{000000} 2.5} & {\color[HTML]{000000} 2.8} \\
{\mtvar} & {\color[HTML]{000000} -2.4} & {\color[HTML]{000000} {\sigi}} & {\color[HTML]{000000} 11.6} & {\color[HTML]{000000} {\sigi}} & {\color[HTML]{B7B7B7} 0.7} & {\color[HTML]{000000} } & {\color[HTML]{B7B7B7} 4.5} & {\color[HTML]{000000} } & {\color[HTML]{000000} 2.3} & {\color[HTML]{000000} 2.9} & {\color[HTML]{000000} 3.0} \\
{\spell} & {\color[HTML]{000000} 8.2} & {\color[HTML]{000000} {\sigiii}} & {\color[HTML]{000000} 68.3} & {\color[HTML]{000000} {\sigiii}} & {\color[HTML]{000000} -2.1} & {\color[HTML]{000000} {\sigi}} & {\color[HTML]{B7B7B7} 8.2} & {\color[HTML]{000000} } & {\color[HTML]{000000} 3.7} & {\color[HTML]{000000} 2.1} & {\color[HTML]{000000} 1.7} \\
{\sttdeu} & {\color[HTML]{B7B7B7} -1.5} & {\color[HTML]{000000} } & {\color[HTML]{B7B7B7} 9.5} & {\color[HTML]{000000} } & {\color[HTML]{000000} 3.2} & {\color[HTML]{000000} {\sigii}} & {\color[HTML]{000000} 10.5} & {\color[HTML]{000000} {\sigi}} & {\color[HTML]{000000} 3.1} & {\color[HTML]{000000} 3.4} & {\color[HTML]{000000} 4.1} \\
{\sttdial} & {\color[HTML]{000000} 4.0} & {\color[HTML]{000000} {\sigiii}} & {\color[HTML]{000000} 17.5} & {\color[HTML]{000000} {\sigii}} & {\color[HTML]{000000} 2.6} & {\color[HTML]{000000} {\sigii}} & {\color[HTML]{000000} 10.7} & {\color[HTML]{000000} {\sigi}} & {\color[HTML]{000000} 4.0} & {\color[HTML]{000000} 3.1} & {\color[HTML]{000000} 3.7} \\
{\tts} & {\color[HTML]{000000} 4.0} & {\color[HTML]{000000} {\sigiii}} & {\color[HTML]{000000} 15.9} & {\color[HTML]{000000} {\sigii}} & {\color[HTML]{B7B7B7} 0.5} & {\color[HTML]{000000} } & {\color[HTML]{B7B7B7} 1.5} & {\color[HTML]{000000} } & {\color[HTML]{000000} 3.8} & 2.9 & 3.1 \\
{\assistantin} & {\color[HTML]{B7B7B7} -0.8} & {\color[HTML]{000000} } & {\color[HTML]{B7B7B7} 7.6} & {\color[HTML]{000000} } & {\color[HTML]{000000} 2.7} & {\color[HTML]{000000} {\sigii}} & {\color[HTML]{B7B7B7} 9.1} & {\color[HTML]{000000} } & {\color[HTML]{000000} 3.5} & 3.7 & 4.2 \\
{\assistantout} & {\color[HTML]{B7B7B7} 1.7} & {\color[HTML]{000000} } & {\color[HTML]{B7B7B7} 3.0} & {\color[HTML]{000000} } & {\color[HTML]{B7B7B7} 0.8} & {\color[HTML]{000000} } & {\color[HTML]{B7B7B7} 2.6} & {\color[HTML]{000000} } & {\color[HTML]{000000} 3.4} & 3.0 & 3.2 \\
{\chatbotin} & {\color[HTML]{B7B7B7} 0.8} & {\color[HTML]{000000} } & {\color[HTML]{B7B7B7} 2.0} & {\color[HTML]{000000} } & {\color[HTML]{B7B7B7} -0.6} & {\color[HTML]{000000} } & {\color[HTML]{B7B7B7} 2.1} & {\color[HTML]{000000} } & {\color[HTML]{000000} 3.4} & 3.2 & 3.1 \\
{\chatbotout} & {\color[HTML]{000000} 3.7} & {\color[HTML]{000000} {\sigiii}} & {\color[HTML]{000000} 13.4} & {\color[HTML]{000000} {\sigii}} & {\color[HTML]{B7B7B7} -0.4} & {\color[HTML]{000000} } & {\color[HTML]{B7B7B7} 2.8} & {\color[HTML]{000000} } & {\color[HTML]{000000} 3.4} & 2.6 & 2.5 \\
{\assistantvar} & {\color[HTML]{000000} -2.3} & {\color[HTML]{000000} {\sigi}} & {\color[HTML]{B7B7B7} 9.2} & {\color[HTML]{000000} } & {\color[HTML]{B7B7B7} 1.1} & {\color[HTML]{000000} } & {\color[HTML]{B7B7B7} 2.1} & {\color[HTML]{000000} } & {\color[HTML]{000000} 2.4} & 2.9 & 3.2 \\
{\search} & {\color[HTML]{000000} 4.8} & {\color[HTML]{000000} {\sigiii}} & {\color[HTML]{000000} 26.1} & {\color[HTML]{000000} {\sigiii}} & {\color[HTML]{B7B7B7} -1.6} & {\color[HTML]{000000} } & {\color[HTML]{B7B7B7} 3.8} & {\color[HTML]{000000} } & {\color[HTML]{000000} 3.8} & 2.8 & 2.4 \\
{\knows} & {\color[HTML]{000000} 4.1} & {\color[HTML]{000000} {\sigiii}} & {\color[HTML]{000000} 13.6} & {\color[HTML]{000000} {\sigiii}} & {\color[HTML]{000000} 8.2} & {\color[HTML]{000000} {\sigiii}} & {\color[HTML]{000000} 49.5} & {\color[HTML]{000000} {\sigiii}} & {\color[HTML]{000000} 0.2} & 0.1 & 0.5 \\
{\appeal} & {\color[HTML]{000000} 5.1} & {\color[HTML]{000000} {\sigiii}} & {\color[HTML]{000000} 35.3} & {\color[HTML]{000000} {\sigiii}} & {\color[HTML]{000000} -3.1} & {\color[HTML]{000000} {\sigii}} & {\color[HTML]{000000} 14.0} & {\color[HTML]{000000} {\sigii}} & {\color[HTML]{000000} 4.3} & 3.3 & 2.7 \\
{\useful} & {\color[HTML]{B7B7B7} 0.1} & {\color[HTML]{000000} } & {\color[HTML]{B7B7B7} 1.4} & {\color[HTML]{000000} } & {\color[HTML]{B7B7B7} 1.6} & {\color[HTML]{000000} } & {\color[HTML]{B7B7B7} 3.8} & {\color[HTML]{000000} } & {\color[HTML]{000000} 3.6} & 3.6 & 3.9 \\
{\often} & {\color[HTML]{000000} 3.7} & {\color[HTML]{000000} {\sigiii}} & {\color[HTML]{000000} 14.8} & {\color[HTML]{000000} {\sigii}} & {\color[HTML]{B7B7B7} -1.6} & {\color[HTML]{000000} } & {\color[HTML]{B7B7B7} 3.1} & {\color[HTML]{000000} } & {\color[HTML]{000000} 3.2} & 2.5 & 2.2\\
\bottomrule
\end{tabular}
\caption{\textbf{Differences between region-based respondent subgroups.}
We show the results of \textit{t-}tests and {\chisq} tests between Low German (NDS) or Swiss (CH) respondents compared to (non-Low-German-speaking) German and Austrian respondents (D/AT).
Positive \textit{t-}statistics indicate that the first group's values for the variable are higher than the second one's, and vice versa for negative values.
\insignificant{Grey entries} denote results with \textit{p-}values $\geq0.05$; asterisks represent smaller \textit{p-}values: {\small*}\,$<0.05$, {\small**}\,$<0.01$, {\small***}\,$<0.001$.
The columns with $\mu$ present the mean Likert scores of the subgroups' responses (e.g., $\mu_{\text{NDS}}$ contains the mean answers provided by our Low Saxon respondents).
The numbers behind the variables refer to the questions in Appendix~\S\ref{sec:full-questionnaire}.
For information on the variables on how the variables are encoded as numbers, see Appendix~\S\ref{sec:correlations}.}
\label{tab:subgroups-region}
\end{table*}

\end{document}